\documentclass[10pt,twocolumn,twoside]{IEEEtran}
\usepackage[american]{babel}
\usepackage{amsfonts,amsmath,latexsym,amssymb,newlfont}
\usepackage{graphicx}
\usepackage[font=footnotesize]{subcaption}
\usepackage{xr-hyper}
\usepackage{acronym}
\usepackage{nccmath}
\usepackage{amsthm}
\usepackage[font=footnotesize]{caption}
\usepackage{subfiles}
\usepackage{algorithm}
\usepackage{multirow} 
\usepackage{float}
\usepackage{cite}
\usepackage{psfrag}
\usepackage{enumitem}
\usepackage{dsfont}
\usepackage{xcolor}
 \usepackage{multicol}
\usepackage{color}
\usepackage[breaklinks=true]{hyperref}
\usepackage[hyphenbreaks]{breakurl}
\usepackage[noend]{algpseudocode}
\usepackage{atveryend}
  \makeatletter
\theoremstyle{definition}
\newtheorem{teorema}{Theorem}

\newtheorem*{lema}{Lemma}

\newcommand\V[1]  { \mathbf{#1} }
\newcommand\B[1]  { \boldsymbol{#1} }
\newcommand\rv[1] {\mathrm{#1}}

\makeatletter
\renewcommand{\fnum@figure}{Fig. \thefigure}
\makeatother
\acrodef{HMM}[HMM]{hidden Markov model}
\acrodef{APLF}[APLF]{adaptive probabilistic load forecasting}
\acrodef{SVM}[SVM]{support vector machine}
\acrodef{LR}[LR]{linear regression}
\acrodef{GP}[GP]{Gaussian process}
\acrodef{QR}[QR]{quantile regression}
\acrodef{SFDA}[SFDA]{secondary forecasting based on deviation analysis}
\acrodef{AFF}[AFF]{adaptive forgetting factor}

\newcommand{\paperTitle}{Probabilistic Load Forecasting Based on Adaptive~Online Learning}
\newcommand{\paperTitleMarkboth}{Probabilistic Load Forecasting Based on Adaptive~Online Learning}

\graphicspath{{images/}{../images/}}

\begin{document}

   \title{\paperTitle}

\author{
	\vspace{0.2cm}
	\IEEEauthorblockN{Ver\'{o}nica \'{A}lvarez,~\IEEEmembership{Student~Member,~IEEE,} 
		Santiago~Mazuelas,~\IEEEmembership{Senior~Member,~IEEE,}  
		 and Jos\'{e}~A.~Lozano,~\IEEEmembership{Fellow,~IEEE}} 	

   \thanks{Manuscript received XXXX, YYYY, and revised AAAA, BBBB.}
   \thanks{
	 This research was supported in part by the Spanish Ministry of Science and Innovation under Ramon y Cajal Grant RYC-2016-19383, Project PID2019-105058GA-I00, and Project TIN2017-82626-R, the BCAM's
Severo Ochoa Excellence Accreditation SEV-2017-0718, the Basque Government through the ELKARTEK and BERC 2018-2021 programmes, Basque Government under the grant  ``Artificial Intelligence in BCAM number EXP. 2019/00432", and the Iberdrola Foundation under the 2019 Research Grants.
         }  
         \thanks{
       V.~\'{A}lvarez is with the BCAM-Basque Center for Applied Mathematics, Bilbao 48009, Spain (e-mail: valvarez@bcamath.org).} 
   \thanks{
       S.~Mazuelas is with the BCAM-Basque Center for Applied Mathematics, and IKERBASQUE-Basque Foundation for Science, Bilbao 48009, Spain (e-mail: smazuelas@bcamath.org).}
          \thanks{
       J.A.~Lozano is with the BCAM-Basque Center for Applied Mathematics, Bilbao 48009, Spain (e-mail: jlozano@bcamath.org).}
}

       \maketitle 

\markboth{IEEE Transactions on Power Systems}{\'{A}lvarez, Mazuelas, and Lozano: \paperTitleMarkboth}

\begin{abstract}
Load forecasting is crucial for multiple energy management tasks such as scheduling generation capacity, planning supply and demand, and minimizing energy trade costs. Such relevance has increased even more in recent years due to the integration of renewable energies, electric cars, and microgrids. Conventional load forecasting techniques obtain single-value load forecasts by exploiting consumption patterns of past load demand. However, such techniques cannot assess intrinsic uncertainties in load demand, and cannot capture dynamic changes in consumption patterns. To address these problems, this paper presents a method for probabilistic load forecasting based on the adaptive online learning of hidden Markov models. We propose learning and forecasting techniques with theoretical guarantees, and experimentally assess their performance in multiple scenarios. In particular, we develop adaptive online learning techniques that update model parameters recursively, and sequential prediction techniques that obtain probabilistic forecasts using the most recent parameters. The performance of the method is evaluated using multiple datasets corresponding with regions that have different sizes and display assorted time-varying consumption patterns. The results show that the proposed method can significantly improve the performance of existing techniques for a wide range of scenarios.
\end{abstract}


\section{Introduction}

\IEEEPARstart{L}{oad forecasting} is crucial for multiple energy management tasks such as scheduling generation capacity, planning supply and demand, and minimizing energy trade costs \cite{plan, planbook, economy, weron:2007}. The importance of load forecasting is growing significantly in recent years due to the increasing development of power systems and smart grids  \cite{weron:2007}. In addition, accurate load forecasting has a beneficial impact in environment and economy by reducing energy waste and purchase~\cite{ranaweera:1997}.

\begin{figure}
\psfrag{Time [Days]}[t][][0.6]{Time [Days]}
\psfrag{Load [GW]}[b][][0.6][180]{Load [GW]}
\psfrag{Load[GW]}[b][][0.6][180]{Load [GW]}
\psfrag{t0}[][][0.6]{$t_0$}
\psfrag{t01}[][][0.6]{$t_0 - 1d$}
\psfrag{t02}[][][0.6]{$t_0 - 2d$}
\psfrag{t03}[][][0.6]{$t_0 - 3d$}
\psfrag{t04}[][][0.6]{$t_0 - 4d$}
\psfrag{t05}[][][0.6]{$t_0 - 5d$}
\psfrag{t06}[][][0.6]{$t_0 - 6d$}
\psfrag{t011}[][][0.6]{$t_0 + 1d$}
\psfrag{t022}[][][0.6]{$t_0 + 2d$}
\psfrag{t033}[][][0.6]{$t_0 + 3d$}
\psfrag{t044}[][][0.6]{$t_0 + 4d$}
\psfrag{t055}[][][0.6]{$t_0 + 5d$}
\psfrag{t066}[][][0.6]{$t_0 + 6d$}
\psfrag{t077}[][][0.6]{$t_0 + 7d$}
\psfrag{t088}[][][0.6]{$t_0 + 8d$}
\psfrag{t099}[][][0.6]{$t_0 + 9d$}
\psfrag{t10}[][][0.6]{$t_0 + 10d$}
\psfrag{t11}[][][0.6]{$t_0 + 11d$}
\psfrag{t12}[][][0.6]{$t_0 + 12d$}
\psfrag{t13}[][][0.6]{$t_0 + 13d$}
\psfrag{t14}[][][0.6]{$t_0 + 14d$}
\psfrag{8}[][][0.6]{8}
\psfrag{11}[][][0.6]{11}
\psfrag{14}[][][0.6]{14}
\psfrag{Offline Learning Predictionnnnnn}[][][0.65]{Offline Learning Prediction}
\psfrag{Online Learning Prediction}[l][l][0.65]{Online Learning Prediction}
\psfrag{Load Demandddddd}[l][l][0.65]{Load Demand}
\includegraphics[width=0.5\textwidth]{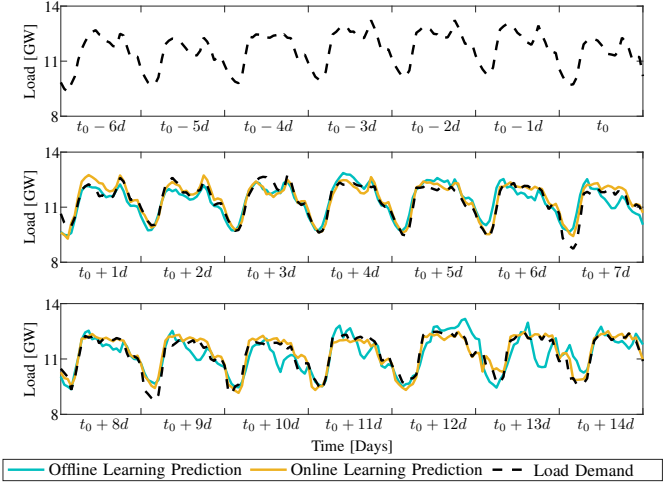}
\caption{The offline learning method cannot capture the change from two-peak to flatter pattern on day $t_0 + 3d$, while the online learning method can harness such change.}
\label{figuraIntro}
\end{figure}

Forecasting methods are enabled by exploiting consumption patterns related to multiple factors such as past loads, hours of day, days of week, holidays, and temperatures \cite{factor, ranaweera:1997, factors, 1033703, narx, weron:2007}.  Accurate forecasting is hindered by intrinsic uncertainties in load demand and dynamic changes in consumption patterns \cite{variability, understanding, impact, drift}. These problems are becoming more relevant in recent years due to the integration of renewable energies, electric cars, and microgrids \cite{micro, micro2, buildingarticle,  technology, cars}. Uncertainties in load demand cannot be assessed by methods that obtain single-value forecasts, and dynamic changes in consumption patterns cannot be captured by methods based on offline learning of static models. On the other hand, probabilistic forecasts can evaluate load uncertainty and are essential for optimal stochastic decision making (e.g., unit commitment \cite{stochastic})  \cite{hong:2016} while online learning is necessary to harness dynamic changes in consumption patterns \cite{online}. Figure~\ref{figuraIntro} illustrates how changes in consumption patterns affect the performance of methods based on offline and online learning. The top figure shows the two-peak consumption pattern that both methods learn on day $t_0$, and the middle figure shows how both methods accurately forecast until a flatter pattern emerges on day $t_0 + 3d$. Then, methods based on offline learning cannot adapt to the new pattern, while methods based on online learning correctly adapt to such change.
 
Most conventional techniques for load forecasting obtain single-value forecasts based on offline learning. Such techniques can be classified in three main groups: techniques based on statistical methods (e.g., \ac{LR} \cite{narx}, autoregressive moving average (ARMA) \cite{pappas:2008, huang:2003}, autoregressive integrated moving average (ARIMA) \cite{amjady:2001}, and seasonal autoregressive integrated moving average (SARIMA) \cite{sarima, sarima1}); techniques based on machine learning methods (e.g., deep learning \cite{deep}, neural networks \cite{hippert:2001}, and \acp{SVM} \cite{svm, narx}); and techniques based on weighted combinations of several forecasts \cite{hybrid, hybrid2, senjyu:2005}. Existing techniques that obtain probabilistic forecasts are based on offline learning, while those based on online learning obtain single-value forecasts. Current probabilistic methods are based on \ac{GP} \cite{gp} and \ac{QR} \cite{quantilRegression, wang2018conditional}. Current online learning methods adapt to dynamic changes in consumption patterns by adjusting offline learning algorithms. In particular, such methods retrain regularly the models of conventional techniques such as ARMA \cite{arma}, update the weights in combined forecasts \cite{combine}, or update the smoothing functions in additive models \cite{adaptivelr}.

In this paper, we present techniques for \ac{APLF} that can harness changes in consumption patterns and assess load uncertainties. Specifically, the main contributions of the paper are as follows:
\begin{itemize}
\item We model the data using \acp{HMM} and develop online learning techniques for \ac{APLF} that update \ac{HMM} parameters recursively.
\item We develop sequential prediction techniques for \ac{APLF} that obtain probabilistic forecasts using the most recent \ac{HMM} parameters.
\item We describe in detail the efficient implementation of the steps for online learning and probabilistic prediction in \ac{APLF} method.
\item We quantify the performance improvement provided by the method presented in comparison with existing load forecasting techniques under multiple scenarios.
\end{itemize}

The rest of this paper is organized as follows. Section \ref{problemFormulation} describes the problem of load forecasting and introduces the performance metrics. In Section \ref{ProposedMethods}, we present the theoretical results for \acsu{APLF} learning and prediction. Section \ref{Imple} compares the procedures of offline learning and online learning, and describes in detail the implementation of \ac{APLF}. The performance of \ac{APLF} and existing techniques is compared in Section \ref{numresults} under multiple scenarios. Finally, Section \ref{concl} draws the conclusions.

\paragraph*{Notations} $N\left( x; \mu, \sigma\right)$ denotes the Gaussian density function of the variable $x$ with mean $\mu$ and standard deviation $\sigma$; $p\left(x|y\right)$ denotes the probability of variable $x$ given variable $y$; $p(x, y)$ denotes the joint probability of variables $x$ and $y$; $\mathds{1} \{\cdot\}$ denotes the indicator function; bold lowercase letters represent vectors; bold capital letters represent matrices; $\V{I}_K$ denotes the $K \times K$ identity matrix; $\V{0}_K$ denotes the zero vector of length $K$; $[\, \cdot \,]$ denotes vectors; and $[\, \cdot \,]^{\text{T}}$ and $\mathbb{E}\{ \cdot \}$ denote the transpose and expectation of its argument.

\section{Problem Formulation}\label{problemFormulation}

Load forecasting methods estimate future loads given past loads and factors that affect future loads such as hours of day, days of week, and weather forecasts. Forecasting techniques determine a prediction function that assigns instance vectors $\V{x}$ (predictors) to target vectors $\V{y}$ (responses).

Instance vectors $\V{x}$ are composed by past loads and observations related to future loads (e.g., weather forecasts), and target vectors $\V{y}$ are composed by future loads. We denote load by $s$ and load forecast by $\hat{s}$, with $s_t$ and $\hat{s}_t$ being the load and the load forecast at time $t$. In addition, for each time $t$, we denote by $\V{r}_t$ the observations vector at time $t$ that can include data such as weather forecasts $w_t$. Then, for a prediction horizon $L$ (e.g., $24$ hours, $30$ minutes) and prediction times $t+1,~t~+~2,~...,~t+L$,  the instance vector is given by $\rv{\V{x}}~=~\left[s_{t-m}, ..., s_{t}, \V{r}_{t+1}^{\text{T}}, ..., \V{r}_{t+L}^{\text{T}}\right]^{\text{T}}$, the target vector is given by $\V{y}~=~[s_{t+1}, s_{t+2}, ..., s_{t+L}]^{\text{T}}$, and the vector of load forecasts is given by $\hat{\V{y}}~=~[\hat{s}_{t+1}, \hat{s}_{t+2}, ..., \hat{s}_{t+L}]^{\text{T}}$. Furthermore, each time $t$ is categorized by a calendar variable $c(t)~\in~\left\{1, 2, ..., C\right\}$ that describes time factors affecting load demand such as hour of day, day of week, month of year, and holiday. The calendar variable is used to model separately loads corresponding with each calendar type $c(t)~\in~\left\{1, 2, ..., C\right\}$ as described in Section \ref{ProposedMethods}. Conventional techniques such as LR \cite{narx} and \acp{SVM} \cite{svm} use instance vectors composed by past loads, observations, and calendar variables. The proposed \ac{APLF} method uses instance vectors composed by one past load and observations. 

Forecasting methods determine prediction functions using training samples formed by pairs of vectors $\V{x}$ and $\V{y}$. Offline learning algorithms determine a static prediction function $f$ using training samples obtained up to time $t_0$, $(\V{x}_1, \V{y}_1), (\V{x}_2, \V{y}_2), ..., (\V{x}_{t_0}, \V{y}_{t_0})$, while online learning algorithms determine prediction functions $f_t$ using all available training samples at $t \geq t_0$, $(\V{x}_1, \V{y}_1), (\V{x}_2, \V{y}_2), ..., (\V{x}_{t}, \V{y}_{t})$. Therefore, the static prediction function $f$ cannot adapt to changes in consumption patterns that occur after time $t_0$, while prediction functions $f_t$ can adapt to patterns' changes using the latest information (see Figure~\ref{figuraIntro}).

Performance of forecasting algorithms is evaluated in terms of accuracy using the absolute value of prediction errors: 
\begin{equation}
\label{predictionerror}
e = \left|s - \hat{s}\right|
\end{equation}
while probabilistic performance can be evaluated using metrics such as pinball losses \cite{dataGEF2014} and calibration \cite{gneiting:2007}. Overall prediction errors are commonly quantified using root mean square error (RMSE) given by
\begin{equation*}
\text{RMSE} = \sqrt{\mathbb{E}\left\{\left|s - \hat{s}\right|^2\right\}}
\end{equation*}
and mean average percentage error (MAPE) given by
\begin{equation*}
\text{MAPE} = 100 \cdot \mathbb{E}\left\{\frac{\left|s - \hat{s}\right|}{s}\right\}.
\end{equation*}
The pinball loss of the $q$-th quantile forecast $\hat{s}^{\left(q\right)}$ is given by
\begin{equation*}
L\big(s, \hat{s}^{\left(q\right)}\big) =  \left\{\begin{matrix}q \left(s - \hat{s}^{\left(q\right)}\right) & \text{if} & s \geq \hat{s}^{\left(q\right)}\\
\left(1-q\right)\left(\hat{s}^{\left(q\right)} - s\right) & \text{if} & s < \hat{s}^{\left(q\right)} \end{matrix}\right.
\label{Pinball}
\end{equation*}
and the overall pinball loss is commonly quantified by the average over all quantiles. The calibration of the $q$-th quantile forecast $\hat{s}^{\left(q\right)}$ is given by
\begin{equation*}
C(q) = \mathbb{E}\left\{\mathds{1}_{s \leq \hat{s}^{\left(q\right)}}\right\}
\end{equation*}
and quantifies the probability with which the load is smaller than the quantile forecast $\hat{s}^{(q)}$. Finally, the expected calibration error (ECE) is given by
\begin{equation*}
\text{ECE} = \mathbb{E}\left\{|q - C(q)|\right\}
\end{equation*}
and quantifies the overall calibration error of probabilistic forecasts.

\section{Models and Theoretical Results}\label{ProposedMethods}

This section first describes the \ac{HMM} that models loads and observations, we then develop the techniques for online learning and probabilistic forecasting.

\begin{figure}
\centering
\psfrag{x1}{{$\rv{\V{r}}_1$}}
\psfrag{x2}{{$\rv{\V{r}}_2$}}
\psfrag{x3}{{$\rv{\V{r}}_{t-1}$}}
\psfrag{xt}{{$\rv{\V{r}}_{t}$}}
\psfrag{y1}{{$s_1$}}
\psfrag{y2}{{$s_2$}}
\psfrag{y3}{{$s_{t-1}$}}
\psfrag{yt}{{$s_t$}}
\psfrag{p1}[][][0.75]{{$p\left(\rv{\V{r}}_1|s_1\right)$}}
\psfrag{p2}[][][0.75]{{$p\left(\rv{\V{r}}_2|s_2\right)$}}
\psfrag{p3}[][][0.75]{}
\psfrag{pt}[][][0.75]{{$p\left(\rv{\V{r}}_t|s_t\right)$}}
\psfrag{q1}[][][0.75]{{$p\left(s_2|s_1\right)$}}
\psfrag{q3}[][][0.75]{{$p\left(s_t|s_{t-1}\right)$}}
\includegraphics[width=0.45\textwidth]{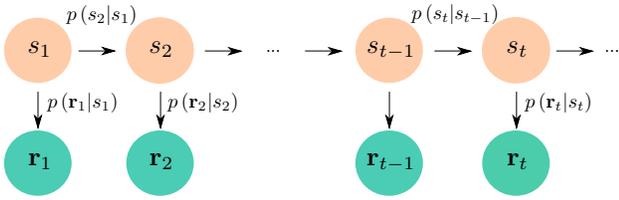}

\vspace{0.3cm}
\caption{Hidden Markov model for sequences $\{s_t\}_{t \geq 1}$ and $\{\V{r}_t\}_{t \geq 1}$ characterized by conditional distributions $p\left(s_t|s_{t-1}\right)$ and $p\left(\rv{\V{r}}_t|s_t\right)$.}
\label{fig: HMMgraph}
\end{figure}

We model the relationship between the loads $\left\{s_t\right\}_{t \geq 1}$ and observations $\left\{\V{r}_t\right\}_{t \geq 1}$ using \acp{HMM} also known as state-space models \cite{hmm, cappe:2006}. Such models allow to predict hidden states from past states and observations, and are determined by the conditional distribution $p\left(s_t|s_{t-1}\right)$ that represents the relationship between two following loads, and the conditional distribution $p\left(\rv{\V{r}}_t|s_t\right)$ that represents the relationship between each load and observations vector (see Figure~\ref{fig: HMMgraph}). We model the sequence of loads and observations as a non-homogeneous \ac{HMM} so that both conditional distributions change in time. Such dynamic modelling allows to adapt to changes in consumption patterns.

The conditional distributions $p\left(s_{t}|s_{t-1}\right)$ and $p\left(\V{r}_t|s_t\right)$ are modeled using Gaussian distributions with mean $\V{u}^{\text{T}} \B{\eta}$ and standard deviation $\sigma$, where $\V{u}$ is a known feature vector and parameters $\B{\eta}$, $\sigma$ are different for each calendar type $c(t) \in \{1, 2, ..., C\}$. For each $c = c(t)$, we denote by $\V{u}_s^{\text{T}} \B{\eta}_{s, c}$ and $\sigma_{s, c}$ the mean and standard deviation that determine the conditional distribution of load at time $t$ given load at time $t-1$ that is
\begin{align}
\label{dgauss}
p\left(s_{t}|s_{t-1}\right) & = N\left(s_t; \V{u}_s^{\text{T}} \B{\eta}_{s, c}, \sigma_{s, c}\right)
\end{align}
with $\B{\eta}_{s, c} \in \mathbb{R}^2$, $\sigma_{s, c} \in \mathbb{R}$, and $\V{u}_s^{\text{T}} = \left[1, s_{t-1}\right]^{\text{T}}$. In addition, for each $c = c(t)$, we denote by $\V{u}_r^{\text{T}} \B{\eta}_{r, c}$ and $\sigma_{r, c}$ the mean and standard deviation that determine the conditional distribution of load at time $t$ given observations at time $t$. Hence, assuming there is no prior knowledge available for the loads, we have that

\begin{align}
\label{rgauss}
p\left(\V{r}_t|s_t\right) & \propto p\left(s_t | \V{r}_t\right) = N\left(s_t; \V{u}_r^{\text{T}} \B{\eta}_{r, c}, \sigma_{r, c}\right)
\end{align}
with $\B{\eta}_{r, c} \in \mathbb{R}^R$, $\sigma_{r, c} \in \mathbb{R}$, and $\V{u}_r = u_r\left(\V{r}_{t}\right) \in \mathbb{R}^R$. The proposed method can consider general functions $u_r(\cdot)$ and observations $\V{r}_t$. In cases where the observations vector is high dimensional, APLF method can use a function $u_r(\cdot)$ that reduces the dimensionality of observations. For instance, if $\V{r} \in \mathbb{R}^N$, $u_r(\V{r}) \in \mathbb{R}^R$ can be the result of applying a dimensionality reduction method such as principal component analysis (PCA). In the experimental results of Section~\ref{numresults}, we use a simple function $u_r(\cdot)$ that returns binary vectors and encodes weather shifts.

Using the above models, at each time $t$ the \ac{HMM} describing the sequences of loads and observations is characterized by parameters
\begin{equation}
\label{thetaparam}
\Theta = \{\B{\eta}_{s, c}, \sigma_{s, c}, \B{\eta}_{r, c}, \sigma_{r, c} : c = 1, 2, ..., C\}
\end{equation}
where $\B{\eta}_{s, c}, \sigma_{s, c}$ characterize the conditional distribution $p(s_t|s_{t-1})$ and $\B{\eta}_{r, c}, \sigma_{r, c}$ characterize the conditional distribution $p(\V{r}_t|s_t)$ for times $t$ with calendar type $c = c(t)$.

The parameters for each calendar type and conditional distribution can be obtained by maximizing the weighted log-likelihood of all the loads obtained at times with the same calendar type. Specifically, if $s_{t_1}, s_{t_2}, ..., s_{t_n}$ are loads obtained at times with calendar type $c \in \{1, 2, ..., C\}$, i.e., $c = c(t_1) = c(t_2) = ... = c(t_n)$, and $\V{u}_{t_1}, \V{u}_{t_2}, ..., \V{u}_{t_n}$ are the corresponding feature vectors for parameters $\B{\eta}_{s, c}, \sigma_{s, c}$ as given in \eqref{dgauss} or for parameters $\B{\eta}_{r, c}, \sigma_{r, c}$ as given in \eqref{rgauss}, the exponentially weighted log-likelihood of loads up to time $t_i$, for $i = 1, 2, ..., n$, is given by 
\begin{equation}
\label{loglike}
L_i\left(\B{\eta}, \sigma\right) = \underset{j = 1}{\overset{i}{\sum}}\lambda^{i-j} \log N(s_{t_j}; \V{u}_{t_j}^{\text{T}} \B{\eta}, \sigma)
\end{equation}
where weights $\lambda^{i-j}$, $j = 1, 2, ..., i$, allow to increase the influence of the most recent data using a parameter  $\lambda \in \left(0, 1\right)$ that is commonly known as forgetting factor. The maximization of \eqref{loglike} is a convex optimization problem since $L_i(\B{\eta}, \sigma)$ is a concave function because Gaussian distributions are log-concave and $\lambda^{i-j} > 0$ for any $i, j$. In addition, the maximum of \eqref{loglike} is unique as long as its Hessian is negative definite which happens for any $i \geq i_0$ such that 
\begin{equation}
\label{nonsmatrix}
\V{H}_{i_0} = \underset{j = 1}{\overset{i_0}{\sum}}\lambda^{{i_0} - j} \V{u}_{t_j} \V{u}_{t_j}^{\text{T}}
\end{equation}
is a non-singular matrix.

The next Theorem shows that the maximization of the weighted log-likelihood in \eqref{loglike} can be solved recursively using parameters given by 
\begin{align}
\label{etarecursion5}
\B{\eta}_i & = \B{\eta}_{i-1} + \frac{\V{P}_{i-1} \V{u}_{t_i}}{\lambda + \V{u}_{t_i}^{\text{T}} \V{P}_{i-1} \V{u}_{t_i}}  \left(s_{t_i} - \V{u}_{t_i}^{\text{T}}\B{\eta}_{i-1} \right)\\
\label{eqsigmarecursion21}
\sigma_{i} & = \sqrt{\sigma_{i-1}^2 -  \frac{1}{\gamma_{i}} \left(\sigma_{i-1}^2-\frac{\lambda^{2}\left(s_{t_i} - \V{u}_{t_i}^{\text{T}}\B{\eta}_{i-1}\right)^2 }{(\lambda + \V{u}_{t_i}^{\text{T}} \V{P}_{i-1} \V{u}_{t_i})^{2}}\right)}
\end{align}
with 
\begin{align}
\label{eqP1}
\V{P}_{i} & = \frac{1}{\lambda}\left(\V{P}_{i-1} -  \frac{\V{P}_{i-1} \V{u}_{t_i}\V{u}_{t_i}^{\text{T}} \V{P}_{i-1}}{\lambda + \V{u}_{t_i}^{\text{T}} \V{P}_{i-1} \V{u}_{t_i}} \right)\\
\label{eqg11}
\gamma_{i} & = 1 + \lambda \gamma_{i-1}.
\end{align}
\begin{teorema} \label{parameters} 
Let $i_0$ be an integer such that the matrix $\V{H}_{i_0}$ given by~\eqref{nonsmatrix} is non-singular, and $\B{\eta}_i^* \in \mathbb{R}^K$, $\sigma_i^* \in \mathbb{R}$ be parameters that maximize the weighted log-likelihood given by \eqref{loglike} with $N(s_{t_j}; \V{u}_{t_j}^{\text{T}} \B{\eta}_i^*, \sigma_i^*) \leq M$ for any $j \leq i \leq n$.

If parameters $\B{\eta}_i \in \mathbb{R}^K$, $\sigma_i  \in \mathbb{R}$ are given by the recursions in \eqref{etarecursion5}-\eqref{eqg11} for $i > 0$ with $\B{\eta}_0 = \V{0}_K$, any $\sigma_0$, $\V{P}_0 = \V{I}_K$, and $\gamma_0 = 0$. Then, we have that
\begin{equation}
\label{aproxerror}
L_i\left(\B{\eta}_i^*, \sigma_i^*\right) - L_i\left(\B{\eta}_i, \sigma_i\right) \leq \pi M^2 \left\| \B{\eta}_i^*\right\|^2  \lambda^i
\end{equation}
for any $i \geq i_0$.

In addition, if parameters $\B{\eta}_i \in \mathbb{R}^K$, $\sigma_i  \in \mathbb{R}$ are given by the recursions in \eqref{etarecursion5}-\eqref{eqg11} for $i > i_0$ with 
\begin{align}
\label{etai0}
\B{\eta}_{i_0} & = \V{P}_{i_0} \Bigg(\underset{j = 1}{\overset{i_0}{\sum}} \lambda^{i_0-j} s_{t_j} \V{u}_{t_j}\Bigg)\\
\label{sigmai0}
\sigma_{i_0} & = \sqrt{\frac{1}{\gamma_{i_0}} \Bigg(\underset{j = 1}{\overset{i_0}{\sum}} \lambda^{{i_0}-j} s_{t_j}^2 - \underset{j = 1}{\overset{i_0}{\sum}} \lambda^{i_0-j} s_{t_j} \V{u}_{t_j}^{\text{T}} \B{\eta}_{i_0}\Bigg)}\\
\label{Pi0}
\V{P}_{i_0} & = (\V{H}_{i_0})^{-1}\\
\label{gammai0}
\gamma_{i_0} & = \underset{j = 1}{\overset{i_0}{\sum}} \lambda^{i_0-j}.
\end{align} 
Then, we have that $\B{\eta}_i = \B{\eta}_i^*$ and $\sigma_i = \sigma_i^*$ for any $i \geq i_0$.

\begin{IEEEproof}
See appendix \ref{app2}.
\end{IEEEproof}

\end{teorema}

The first part of the above result shows that parameters given by the recursions \eqref{etarecursion5}-\eqref{eqg11} for $i>0$ with $\B{\eta}_0 = \V{0}_K$, any $\sigma_0$, $\V{P}_0 = \V{I}_K$, and $\gamma_0 = 0$, approximately maximize the weighted log-likelihood. In addition, the log-likelihood difference with respect to the maximum given by \eqref{aproxerror} decreases exponentially fast with the number of iterations $i$ since $\lambda \in (0, 1)$. The second part shows that parameters given by the recursions \eqref{etarecursion5}-\eqref{eqg11} for $i \geq i_0$ with $\B{\eta}_{i_0}, \sigma_{i_0}$ given by~\eqref{etai0}-\eqref{gammai0}, maximize the weighted log-likelihood for any $i \geq i_0$. Note that the parameters are updated in recursions \eqref{etarecursion5}-~\eqref{eqg11} by adding a correction to the previous parameters $\B{\eta}_{i-1}$ and $\sigma_{i-1}$. Such correction is proportional to the fitting error of the previous parameter $s_{t_i} - \V{u}_{t_i}^{\text{T}} \B{\eta}_{i-1}$ so that parameters are updated depending how well they fit the most recent data. 

Recursion \eqref{etarecursion5} for parameters describing means is similar to that used for the recursive minimization of weighted least squares \cite{onlineregression}. The main technical novelty in Theorem~\ref{parameters} lies in recursion \eqref{eqsigmarecursion21} for parameters describing standard deviations, and inequality \eqref{aproxerror} describing the near-optimality of parameters initialized with $\B{\eta}_0 = \V{0}_K$, any $\sigma_0$, $\V{P}_0 = \V{I}_K$, and $\gamma_0 = 0$. Existing techniques for least squares only allow to recursively obtain parameters describing means, while Theorem~\ref{parameters} allows to recursively obtain parameters describing both means and standard deviations. Such generalization is of practical relevance since it allows to obtain probabilistic models determined by time-changing means and standard deviations. In addition, existing techniques address the possible singularity of matrix \eqref{nonsmatrix} during the initial steps by adding a regularization term in \eqref{loglike}, but such approach cannot be used to obtain standard deviations. The bound in \eqref{aproxerror} guarantees that parameters given by recursions \eqref{etarecursion5}-\eqref{eqg11} are close to be optimal when initialized with $\B{\eta}_0 = \V{0}_K$, any $\sigma_0$, $\V{P}_0 = \V{I}_K$ and $\gamma_0 = 0$, and are optimal when initialized as given by~\eqref{etai0}-\eqref{gammai0}.

\begin{figure}
\centering
{
\psfrag{4 std}[l][l][0.65]{{$\hat{s}_t \pm 4\hat{e}_t$}}
\psfrag{2 std}[l][l][0.65]{{$\hat{s}_t \pm 2\hat{e}_t$}}
\psfrag{Load forecast 555555}[l][l][0.65]{{Load forecast $\hat{s}_t$}}
\psfrag{Load}[l][l][0.65]{{Load $s_t$}}
\psfrag{9}{{}}
\psfrag{11}{{}}
\psfrag{13}{{}}
\psfrag{1}[][][0.6]{1}
\psfrag{2}[][][0.6]{2}
\psfrag{3}[][][0.6]{3}
\psfrag{4}[][][0.6]{4}
\psfrag{5}[][][0.6]{5}
\psfrag{6}[][][0.6]{6}
\psfrag{7}[][][0.6]{7}
\psfrag{8}[][][0.6]{8}
\psfrag{10}[][][0.6]{10}
\psfrag{12}[][][0.6]{12}
\psfrag{14}[][][0.6]{14}
\psfrag{Load [GW]}[b][][0.7]{{Load [GW]}}
\psfrag{Time [Days]}[t][][0.7]{{Time [Days]}}
\includegraphics[width=0.5\textwidth]{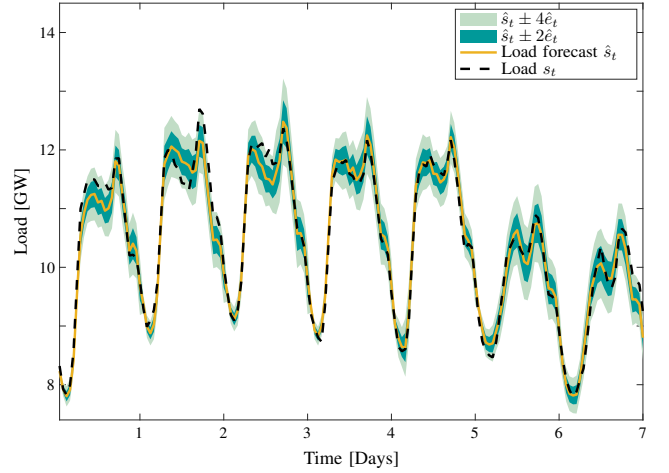}
}
\caption{APLF method obtains load forecasts together with reliable uncertainty assessments of load demand.}
\label{figst}
\end{figure}

\begin{figure*}
\centering
\begin{subfigure}[t]{1\textwidth}
\psfrag{x1}[][][0.6]{$\V{x}_1$}
\psfrag{y1}[][][0.6]{$\V{y}_1$}
\psfrag{x2}[][][0.6]{$\V{x}_2$}
\psfrag{y2}[][][0.6]{$\V{y}_2$}
\psfrag{xn}[][][0.6]{$\V{x}_{t_0}$}
\psfrag{yn}[][][0.6]{$\V{y}_{t_0}$}
\psfrag{x8}[][][0.6]{$\V{x}_{{t_0}+1}$}
\psfrag{y8}[][][0.6]{$\V{y}_{{t_0}+1}$}
\psfrag{x9}[][][0.6]{$\V{x}_{{t_0}+2}$}
\psfrag{y9}[][][0.6]{$\V{y}_{{t_0}+2}$}
\psfrag{x8p}[][][0.55]{$\hat{\V{y}}_{{t_0}+1}$}
\psfrag{x9p}[][][0.55]{$\hat{\V{y}}_{{t_0}+2}$}
\psfrag{H}{{}}
\includegraphics[width=1\textwidth]{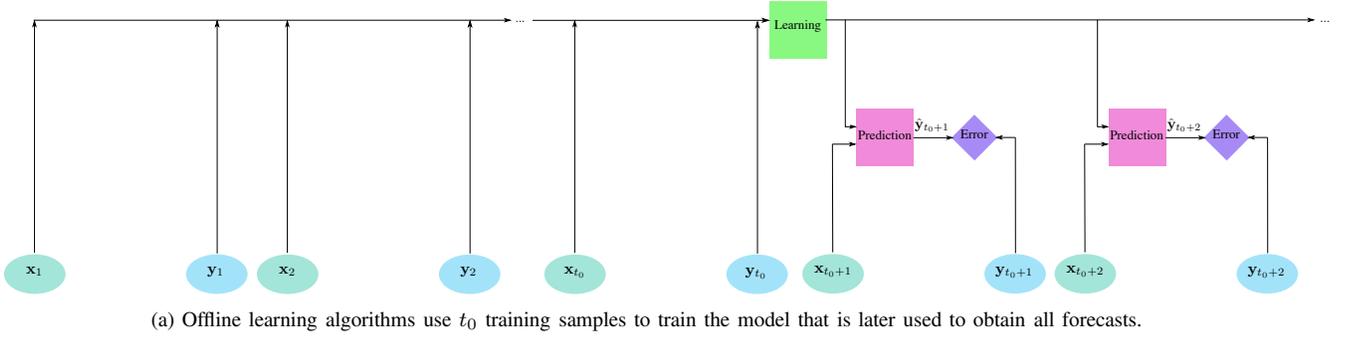}
\caption{Offline learning algorithms use $t_0$ training samples to train the model that is later used to obtain all forecasts.}
\label{procoffline}
\end{subfigure}\\
\vspace{0.7cm}
\begin{subfigure}[t]{1\textwidth}
\centering
\psfrag{x1}[][][0.6]{{$\V{x}_1$}}
\psfrag{y1}[][][0.6]{{$\V{y}_1$}}
\psfrag{x2}[][][0.6]{{$\V{x}_2$}}
\psfrag{y2}[][][0.6]{{$\V{y}_2$}}
\psfrag{xn}[][][0.6]{{$\V{x}_{t_0}$}}
\psfrag{yn}[][][0.6]{{$\V{y}_{t_0}$}}
\psfrag{x8}[][][0.6]{{$\V{x}_{{t_0}+1}$}}
\psfrag{y8}[][][0.6]{{$\V{y}_{{t_0}+1}$}}
\psfrag{x9}[][][0.6]{{$\V{x}_{{t_0}+2}$}}
\psfrag{y9}[][][0.6]{{$\V{y}_{{t_0}+2}$}}
\psfrag{x1p}[][][0.55]{$\hat{\V{y}}_1$}
\psfrag{x2p}[][][0.55]{$\hat{\V{y}}_2$}
\psfrag{xnp}[][][0.55]{$\hat{\V{y}}_{t_0}$}
\psfrag{x8p}[][][0.55]{$\hat{\V{y}}_{{t_0}+1}$}
\psfrag{x9p}[][][0.55]{$\hat{\V{y}}_{{t_0}+2}$}
\psfrag{y1p}[][][0.55]{$\hat{\V{e}}_1$}
\psfrag{y2p}[][][0.55]{$\hat{\V{e}}_2$}
\psfrag{ynp}[][][0.55]{$\hat{\V{e}}_{t_0}$}
\psfrag{y8p}[][][0.55]{$\hat{\V{e}}_{{t_0}+1}$}
\psfrag{y9p}[][][0.55]{$\hat{\V{e}}_{{t_0}+2}$}
\includegraphics[width=1\textwidth]{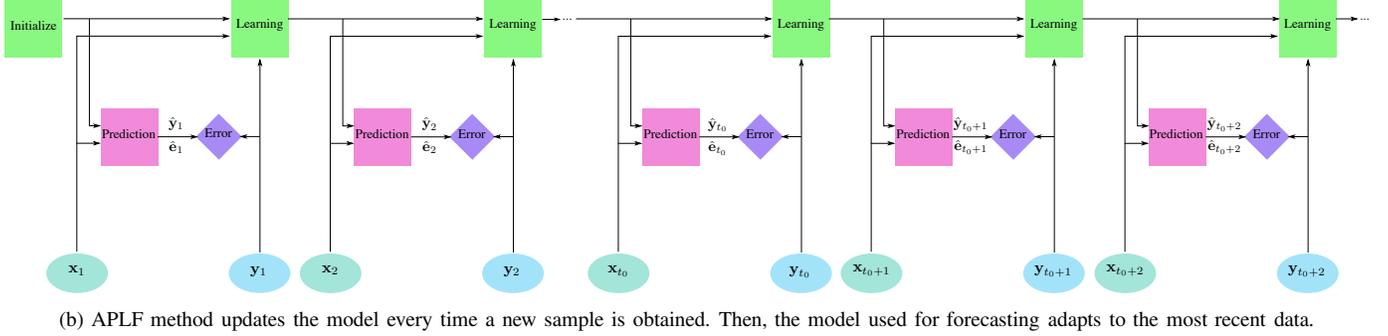}
\caption{\ac{APLF} method updates the model every time a new sample is obtained. Then, the model used for forecasting adapts to the most recent data.}
\label{proconline}
\end{subfigure}
\caption{Block diagrams for offline learning algorithms and \ac{APLF} method.}
\label{block}
\end{figure*}

The above Theorem enables the adaptive online learning of parameters $\Theta$ described in \eqref{thetaparam}. As detailed in Section~\ref{Imple}, Theorem~\ref{parameters} allows to update parameters $\B{\eta}_{s, c}$, $\sigma_{s, c}$ and $\B{\eta}_{r, c}$, $\sigma_{r, c}$ using the recursions \eqref{etarecursion5} and \eqref{eqsigmarecursion21} every time new loads and observations are obtained corresponding with calendar type $c$. Such parameters are updated using their previous values and the states variables given by \eqref{eqP1} and \eqref{eqg11}. In the following, we denote $\V{P}_{s, c}$ and $\gamma_{s, c}$ (resp. $\V{P}_{r, c}$ and $\gamma_{r, c}$) the state variables required to update parameters $\B{\eta}_{s, c}$ and $\sigma_{s, c}$ (resp. $\B{\eta}_{r, c}$ and $\sigma_{r, c}$) for $c = 1, 2, ..., C$, and we denote by $\Gamma$ the list composed by those state variables, that is
\begin{equation}
\label{Gammaparam}
\Gamma = \{\V{P}_{s, c}, \gamma_{s, c}, \V{P}_{r, c}, \gamma_{r, c} : c = 1, 2, ..., C\}.
\end{equation}
 In addition, we denote $\lambda_s$ (resp. $\lambda_r$)  the forgetting factor required to update parameters $\B{\eta}_{s, c}$ and $\sigma_{s, c}$ (resp. $\B{\eta}_{r, c}$ and $\sigma_{r, c}$), for $c = 1, 2, ..., C$. 
 
The previous result describes how to update HMM parameters using the most recent data, the next result shows how to obtain probabilistic forecasts using the \ac{HMM} characterized by parameters $\Theta$.

\begin{teorema}\label{teoremaRecursion}
If $\left\{s_t, \V{r}_t\right\}_{t \geq 1}$ is an \acsu{HMM} characterized by parameters $\Theta$ as described in \eqref{thetaparam}. Then, for $i = 1, 2, ..., L$
\begin{equation}
\label{pgauss}
p\left(s_{t+i}|s_t, \V{r}_{t + 1}, ..., \V{r}_{t +i}\right) = N\left(s_{t+i}; \hat{s}_{t+i}, \hat{e}_{t+i}\right)
\end{equation}
where means $\hat{s}_{t+1}, \hat{s}_{t+2}, ..., \hat{s}_{t+L}$ and standard deviations $\hat{e}_{t+1}, \hat{e}_{t+2}, ..., \hat{e}_{t+L}$ can be computed by the following recursions
\begin{align}
c & = c\left(t+i\right), \hat{\V{u}}_s = [1, \hat{s}_{t+i-1}]^{\text{T}}, \V{u}_r = u_r\left({\V{r}}_{t+i}\right) \nonumber\\
\label{spred}
\hat{s}_{t+i} & = \frac{\hat{\V{u}}_s^{\text{T}} \B{\eta}_{s, c} \sigma_{r, c}^2 + \V{u}_r^{\text{T}} \B{\eta}_{r, c} \left(\sigma_{s, c}^2 + \left(\V{v}^{\text{T}} \B{\eta}_{s, c}\right)^2\hat{e}_{t+i-1}^2\right)}{\sigma_{r, c}^2 + \sigma_{s, c}^2 + \left(\V{v}^{\text{T}} \B{\eta}_{s, c}\right)^2\hat{e}_{t+i-1}^2}\\
\label{epred}
\hat{e}_{t+i} & = \sqrt{\frac{\sigma_{r, c}^2 \left( \sigma_{s, c}^2 +\left(\V{v}^{\text{T}} \B{\eta}_{s, c}\right)^2\hat{e}_{t+i-1}^2\right)}{\sigma_{r, c}^2 + \sigma_{s, c}^2 + \left(\V{v}^{\text{T}}\B{\eta}_{s, c}\right)^2\hat{e}_{t+i-1}^2}}
\end{align}
for $\V{v} = \left[0, 1\right]^{\text{T}}$, $\hat{s}_t = s_t$, $\hat{e}_t = 0$, and $i = 1, 2, ..., L$.

\begin{IEEEproof}
See appendix \ref{ap3}.
\end{IEEEproof}

\end{teorema}

The above Theorem enables to recursively obtain probabilistic load forecasts $N\left(s_{t+i}; \hat{s}_{t+i}, \hat{e}_{t+i}\right)$, for $i = 1, 2, ..., L$ that allow to quantify the probability of forecast intervals (see Figure~\ref{figst}). As detailed in Section~\ref{Imple}, Theorem~\ref{teoremaRecursion} allows to obtain load forecasts $\hat{s}_{t+i}$ together with estimates of their accuracy $\hat{e}_{t+i}$ for each $i = 1, 2, ..., L$. Such forecasts are obtained using the recursions \eqref{spred} and \eqref{epred} with the most recent parameters every time new instance vectors $\V{x}$ are obtained. Specifically, for each $i = 1, 2, ..., L$, the probabilistic forecast at time $t+i$, $N\left(s_{t+i}; \hat{s}_{t+i}, \hat{e}_{t+i}\right)$, is obtained using 1) the probabilistic forecast $N\left(s_{t+i-1}; \hat{s}_{t+i-1}, \hat{e}_{t+i-1}\right)$ at previous time $t+i-1$; 2) observations vector $\V{r}_{t+i}$ at time $t+i$; and 3) parameters $\B{\eta}_{s, c}, \sigma_{s, c}$ and $\B{\eta}_{r, c}, \sigma_{r, c}$ corresponding with calendar type at time $t+i$, $c = c(t+i)$. 

The results in this section provide theoretical guarantees for the learning and prediction steps of \ac{APLF} method. The next section describes \ac{APLF} in comparison with approaches based on offline learning, and details the implementation steps for learning and prediction using APLF.

\section{Implementation}\label{Imple}

Offline learning methods obtain one model, while online learning methods obtain a sequence of models. In particular, \ac{APLF} method learns a model every time a new sample is obtained. Figure \ref{block} describes the block diagrams for load forecasting based on offline learning and based on \ac{APLF} method.

Load forecasting methods based on offline learning train a model using a set of samples. Later, such methods predict the load for each new instance vector using the learned model. At learning, $t_0$ training samples are used to obtain the model following different approaches. For instance, ARMA algorithm calculates parameters for the autoregressive, moving average, and error terms \cite{pappas:2008}, while techniques based on machine learning calculate parameters determining a regression function \cite{narx, svm,  fs}. At prediction, the learned model and the instance vector $\V{x}_t$ are used to obtain load forecasts $\hat{\V{y}}_t$ at time $t$, for $t>{t_0}$. The model used in the prediction step is the same for all times $t$, and the actual loads $\V{y}_t$ for $t > t_0$ are only used to quantify the prediction error (see Figure~\ref{procoffline}).

Load forecasting methods based on online learning train models regularly using the most recent samples. Later, such methods predict for each new instance vector using the latest learned model. At learning, training samples and possibly the previous model are used to obtain a new model following different approaches. For instance, the method proposed in \cite{arma} recalculates parameters of the ARMA algorithm, the method proposed in \cite{combine} recalculates the weights in combined forecasts, and the method proposed in \cite{adaptivelr} updates the smoothing functions in additive models. At prediction, the latest learned model and the instance vector $\V{x}_t$ are used to obtain load forecasts $\hat{\V{y}}_t$ at time $t$. 

\ac{APLF} is a forecasting method based on online learning that updates model parameters using recursions in Theorem~\ref{parameters} and obtain probabilistic forecasts as given by Theorem~\ref{teoremaRecursion}. At learning, \ac{APLF} obtains the new model using the instance vector $\V{x}_{t}$, the actual loads $\V{y}_t$, and the previous model. At prediction, \ac{APLF} uses the latest model to obtain load forecasts $\hat{\V{y}}_t = \left[\hat{s}_{t+1}, \hat{s}_{t+2}, ..., \hat{s}_{t+L}\right]^{\text{T}}$ and estimated errors $\hat{\V{e}}_t = \left[\hat{e}_{t+1}, \hat{e}_{t+2}, ...,\hat{e}_{t+L}\right]^{\text{T}}$ that determine probabilistic forecasts as given in \eqref{pgauss}. The model used in the prediction step adapts at each time $t$ to the most recent data, and the actual loads $\V{y}_t$ are not only used to quantify the error, but also to update the model (see Figure~\ref{proconline}). 

Algorithm~\ref{alg1} and Algorithm~\ref{alg2} detail the efficient implementation of the learning and prediction steps of APLF. The corresponding source code in Python and Matlab languages is publicly available on the web \url{https://github.com/MachineLearningBCAM/Load-forecasting-IEEE-TPWRS-2020}. The running times of Algorithm~\ref{alg1} and Algorithm~\ref{alg2} are amenable for real-time implementation with very low latency since APLF has memory complexity $O(CR^2)$, the learning step has computational complexity $O(L R^3)$, and the prediction step has computational complexity $O(L R)$. Note that the values of $R$ are small in practice, for instance we use $R = 3$, $L = 24$, and $C = 48$ in the numerical results of Section~\ref{numresults}. Algorithm~\ref{alg1} follows recursions given in Theorem~\ref{parameters} for parameters $\Theta$ and state variables $\Gamma$ described in \eqref{thetaparam} and \eqref{Gammaparam}, respectively. Specifically, such algorithm updates parameters $\B{\eta}_{s, c}, \sigma_{s, c}$ and $\B{\eta}_{r, c}, \sigma_{r, c}$ as well as state variables $\V{P}_{s, c}, \gamma_{s, c}$ and $\V{P}_{r, c}, \gamma_{r, c}$ using instances and actual loads with calendar type $c$. Algorithm~\ref{alg2} follows recursions given in Theorem~\ref{teoremaRecursion} using the parameters $\Theta$ described in \eqref{thetaparam} and the new instance vector. Specifically, such algorithm obtains $L$ load forecasts and $L$ estimates of their accuracy using the latest parameters $\B{\eta}_{s, c}, \sigma_{s, c}$ and $\B{\eta}_{r, c}, \sigma_{r, c}$ and the corresponding instance vector. Note that the proposed method can predict at any time of the day and use general prediction horizons $L$. In addition, these prediction times and horizons can change from one day to another just by modifying the corresponding inputs in Algorithm~\ref{alg2}.
 \begin{algorithm}
\small
\caption{Learning step for APLF}
\label{alg1}
\begin{algorithmic}
\State \textbf{Input:} \hspace{0.3cm}$\Theta$ \hspace{4.1cm}model parameters\\ 
\hspace{1.28cm}$\Gamma$ \hspace{4.15cm}state variables\\ 
\hspace{1.28cm}$\lambda_s$, $\lambda_r$ \hspace{3.47cm}forgetting factors\\ \hspace{1.28cm}$\V{x}_t=\left[s_t, \V{r}_{t+1}^{\text{T}}, \V{r}_{t+2}^{\text{T}}, ..., \V{r}_{t+L}^{\text{T}}\right]^{\text{T}}$ \hspace{0.22cm}new instance vector\\
\hspace{1.28cm}$\V{y}_t = \left[s_{t+1}, s_{t+2}, ..., s_{t+L}\right]^{\text{T}}$ \hspace{0.72cm}new loads vector\\ 
\hspace{1.2cm} $t$ \hspace{4.2cm}time\\
\textbf{Output:}\hspace{0.18cm}$\Theta$ \hspace{0.34cm}updated model parameters\\ 
\hspace{1.28cm}$\Gamma$ \hspace{0.37cm}updated state variables
\vspace{0.2cm}
		\For{$i = 1, 2, ..., L$} 
		\vspace{0.1cm}
				\State
		$c \gets c\left(t+i\right)$
		\vspace{0.1cm}
             \State $\V{u}_s \gets \left[1, s_{t+i-1}\right]^{\text{T}}$
             \vspace{0.1cm}
                      \State $\V{u}_r \gets u_r\left(\V{r}_{t+i}\right)$
                      \vspace{0.1cm}
                      \For{$j = s, r$}
                          \begin{align*}
          \V{P}_{j, c} & \gets \frac{1}{\lambda_j}\left(\V{P}_{j, c} -  \frac{\V{P}_{j, c} \V{u}_j \V{u}_j^{\text{T}} \V{P}_{j, c}}{\lambda_j + \V{u}_j^{\text{T}} \V{P}_{j, c} \V{u}_j} \right)\\
         \gamma_{j, c} & \gets 1 + \lambda_j \gamma_{j, c}\\
         \sigma_{j, c} & \gets \sqrt{\sigma_{j, c}^2 - \frac{1}{\gamma_{j, c}} \left(\sigma_{j, c}^2-\frac{\lambda_j^{2}\left(s_{t+i} - \V{u}_j^{\text{T}} \B{\eta}_{j, c}\right)^2 }{(\lambda_j + \V{u}_j^{\text{T}} \V{P}_{j, c} \V{u}_j)^{2}}\right)}\\
           \B{\eta}_{j, c} & \gets \B{\eta}_{j, c} + \frac{\V{P}_{j, c} \V{u}_j}{\lambda_j + \V{u}_j^{\text{T}} \V{P}_{j, c} \V{u}_j} \left(s_{t+i} - \V{u}_j^{\text{T}} \B{\eta}_{j, c}\right)
           \end{align*}
      	\EndFor
	\EndFor
\end{algorithmic}
\end{algorithm}

\color{black}
\begin{algorithm}
\color{black}
\small
\caption{Prediction step for APLF}
\label{alg2}
\begin{algorithmic}
\State \textbf{Input:} \hspace{0.32cm}$\Theta$ \hspace{4.3cm}model parameters\\ 
\hspace{1.2cm} $\V{x}_t = \left[s_t, \V{r}_{t+1}^{\text{T}}, \V{r}_{t+2}^{\text{T}}, ..., \V{r}_{t+L}^{\text{T}}\right]^{\text{T}}$ \hspace{0.4cm}new instance vector\\
\hspace{1.2cm} $t$ \hspace{04.42cm}time\\
\textbf{Output:}\hspace{0.2cm}$\hat{\V{y}}_t = \left[\hat{s}_{t+1}, \hat{s}_{t+2}, ..., \hat{s}_{t+L}\right]^{\text{T}}$ \hspace{0.94cm}load forecasts \\
\hspace{1.31cm}$\hat{\V{e}}_t = \left[\hat{e}_{t+1}, \hat{e}_{t+2}, ..., \hat{e}_{t+L}\right]^{\text{T}}$ \hspace{0.95cm}estimated errors \\
\hspace{1.31cm}$N(s_{t+i}; \hat{s}_{t+i}, \hat{e}_{t+i})$, $i = 1, 2, ..., L$ \hspace{-0.07cm} prob. forecasts
\State$\hat{s}_t \gets s_t$ \\
$\hat{e}_t = 0$
\vspace{0.1cm}
		\For{$i = 1, 2, ..., L$} 
		\vspace{0.1cm}
		\State 
           $c \gets c\left(t+i\right)$
           \vspace{0.1cm}
            \State $\hat{\V{u}}_s \gets \left[1, \hat{s}_{t+i-1}\right]^{\text{T}}$
            \vspace{0.1cm}
                  \State $\V{u}_r \gets u_r\left(\V{r}_{t+i}\right)$
                  \vspace{0.1cm}
            \State Obtain load forecast $\hat{s}_{t+i}$ using equation \eqref{spred}
            \vspace{0.1cm}
            \State Obtain prediction error $\hat{e}_{t+i}$ using equation \eqref{epred}
		\EndFor
\end{algorithmic}
\end{algorithm}

\section{Numerical results}\label{numresults}

This section first describes the datasets used for the experimentation, and then compares the performance of \ac{APLF} method with respect to that of existing techniques. The first set of numerical results quantifies the prediction errors, while the second set of numerical results evaluates the performance of probabilistic load forecasts and analyzes the relationship between training size and prediction error. 

\begin{figure}
        \psfrag{Time [Days of year]}[b][][0.8]{Time [Days of year]}
        \psfrag{Time [Hours of day]}[t][][0.8]{Time [Hours of day]}
                 \psfrag{dayton1}[b][][0.8]{Dayton 2004}
                  \psfrag{dayton2}[b][][0.8]{Dayton 2005}
        \psfrag{[GW]}[t][][0.8]{{[GW]}}
        \psfrag{6}[][][0.6]{6}
\psfrag{12}[][][0.6]{12}
\psfrag{18}[][][0.6]{18}
\psfrag{24}[][][0.6]{24}
\psfrag{1}[][][0.6]{1}
\psfrag{1.5}[][][0.6]{1.5}
\psfrag{2}[][][0.6]{2}
\psfrag{2.5}[][][0.6]{2.5}
\psfrag{3}[][][0.6]{3}
\psfrag{3.5}[][][0.6]{3.5}
\psfrag{50}[][][0.6]{50}
\psfrag{100}[][][0.6]{100}
\psfrag{150}[][][0.6]{150}
\psfrag{200}[][][0.6]{200}
\psfrag{250}[][][0.6]{250}
\psfrag{300}[][][0.6]{300}
\psfrag{350}[][][0.6]{350}  
\vspace{-0cm}\includegraphics[width=0.5\textwidth]{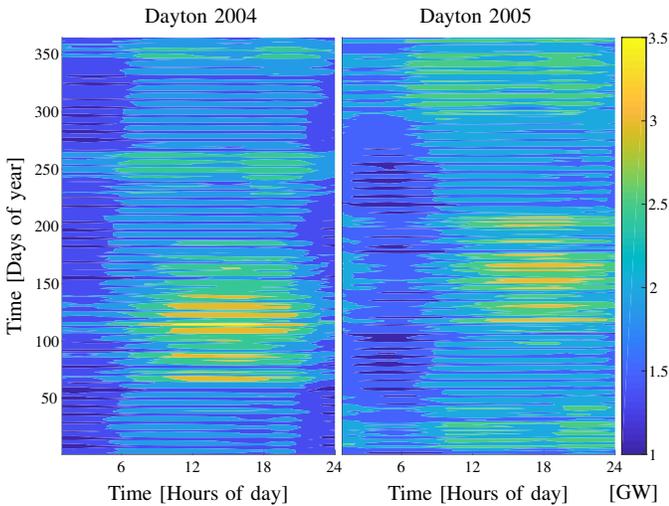}
        \caption{Consumption patterns can significantly change in two consecutive years.}
        \label{differUSA}
\end{figure}

\begin{table*}
\centering
\small
\caption{RMSE and MAPE of prediction errors for \ac{APLF} and $11$ state-of-the-art techniques on $7$ datasets.}
\renewcommand{\arraystretch}{1.2}
\begin{tabular}{|c|rrrr|rrrrrr|rrrr|}
\hline
\multirow{3}{*}{Method} & \multicolumn{4}{c|}{Large-size region} & \multicolumn{6}{c|}{Medium-size region}           & \multicolumn{4}{c|}{Small-size region}                  \\ 
                               & \multicolumn{2}{c}{Belgium}  & \multicolumn{2}{c|}{New Engld.}  & \multicolumn{2}{c}{GEFCom12} & \multicolumn{2}{c}{GEFCom2014} &  \multicolumn{2}{c|}{Dayton}  & \multicolumn{2}{c}{400 builds.} & \multicolumn{2}{c|}{100 builds.}              \\ 
                                   & [GW] & [$\%$] & [GW]  & [$\%$] & [MW]   &[$\%$]  &  [MW] & [$\%$]  & [GW]  & [$\%$]  & [kW] & [$\%$]  &     [kW] & [$\%$]  \\ \hline                
LR                     & $1.47$ & $11.8$          & $1.73$    &  $8.0$ & $5.48$ &  $20.6$  & $0.30$   &  $15.0$ & $0.46$ & $15.5$ & $0.06$   &   $9.1$  & $0.07$     &                $14.3$ \\ 
SARIMA              &$0.81$  &  $5.5$            & $1.22$  &  $5.4$ &$3.42$   & $11.6$ &  $0.25$  & $12.0$ & $0.22$    & $7.7$ &$0.05$    &  $9.9$   &             $0.07$& $16.5$ \\ 
QR & $1.05$ & $9.2$ & $1.17$ & $5.6$ & $5.46$ & $25.5$ & $0.29$ & $14.7$ & $0.20$ & $7.1$   & $0.08$ & $17.6$ & $0.12$ & $26.8$ \\ 
GP & $\textbf{0.52}$ & $\textbf{4.0}$ & $0.89$ & $4.2$ & $2.50$ & $\textbf{8.5}$ & $\textbf{0.24}$ & $\textbf{10.6}$ & $0.19$ & $6.3$   & $\textbf{0.04}$& $\textbf{6.8}$  &$\textbf{0.05}$ & $\textbf{11.2}$ \\ 
\ac{SVM}                   & $0.69$ & $4.7$       &    $1.11$  & $5.5$  &   $3.28$     & $12.2$  &         $\textbf{0.24}$  &  $12.3$ &     $\textbf{0.17}$    & $\textbf{5.7}$&  $\textbf{0.04}$   & $8.0$ &   $\textbf{0.05}$    &  $11.4$\\
DRN             &  $1.74$ & $13.0$                 & $\textbf{0.52}$ & $\textbf{2.3}$ & $\textbf{2.17}$  & $7.6$  & $0.27$ & $19.5$ & $0.31$ & $11.3$& $\textbf{0.04}$    & $7.0$   & $0.07$ & $14.7$ \\
AR                & $0.66$  &  $5.1$          &  $1.28$      & $5.6$  &   $3.94$      &        $16.2$ &    $0.30$          & $18.5$ &        $0.38$   &   $13.6$ &      $\textbf{0.04}$ &   $8.9$   &      $0.07$      & $16.9$  \\
ARNFS           & $1.08$ & $9.4$     &    $1.95$   &  $10.9$ &     $4.41$   &   $17.4$  &      $0.33$  &  $18.4$   &     $0.31$      &    $14.1$ & $\textbf{0.04}$ &   $8.1$  &  $\textbf{0.05}$ & $11.4$ \\ 
ARRFFS       &  $1.18$ & $10.3$     & $2.05$   &   $11.2$ &      $4.54$  &  $17.2$ &  $0.34$    & $17.7$ &      $0.28$     &  $10.3$  & $0.08$  &  $17.5$   &   $0.10$ & $23.4$ \\
SFDA            &$ {1.14}$  & $8.9$       & $1.41$    &  $10.2$ & ${5.04}$  &  $16.8$ & ${0.35}$  & $14.5$  & $0.39$  & ${21.8}$  & {0.06}   &{13.0}&{0.08}  & {18.6} \\
{{AFF}}             &$  {0.95}$  & ${6.7} $       & $ {1.23}$    &  ${5.8}$ & ${2.91}$  &  ${10.7}$ & ${0.25}$  & ${15.2}$  & ${0.26}$  & ${9.6}$  &  {0.05}   &  {10.2}&{0.07}  &{14.6} \\
 \ac{APLF}                   &$\textbf{0.33}$  & $\textbf{2.3}$       & $\textbf{0.86}$    &  $\textbf{3.9}$ & $\textbf{2.15}$  &  $\textbf{8.1}$ & $\textbf{0.20}$  & $\textbf{9.6}$  & $\textbf{0.16}$  & $\textbf{5.5}$  & \textbf{0.03}    & \textbf{6.3}   & \textbf{0.05}    &  \textbf{11.0} \\ \cline{1-15} 
\end{tabular}
\label{tabla1}
\end{table*}

Seven publicly available datasets are selected for numerical experimentation. The datasets correspond with regions that have different sizes and display different consumption patterns that change over time. Such changes can be observed in Figure~\ref{differUSA} that shows load demand per hour of day and day of year during 2004 and 2005 in Dayton (US). This figure shows that consumption patterns change significantly not only for different seasons but also between consecutive weeks and between consecutive years. Therefore, methods based on static models often obtain inferior accuracies since they cannot adapt to dynamic changes in consumption patterns. 

\begin{figure*}
    \centering
     \begin{subfigure}[t]{0.33\textwidth}
        \centering
                \psfrag{0.9}{{}}
\psfrag{Load [GW]}[b][][0.6]{{Load [GW]}}
        \psfrag{Time [Hours]}[t][][0.6]{{Time [Hours]}}
                \psfrag{Steps ahead forecast}[b][][0.6]{{Prediction horizon}}
\psfrag{HMM}[l][l][0.45]{APLF}
\psfrag{SVM}[l][l][0.45]{SVM}
\psfrag{QR}[l][l][0.45]{QR}
\psfrag{SARIMAAAAA}[l][l][0.45]{SARIMA}
\psfrag{AR}[l][l][0.45]{AR}
\psfrag{LR}[l][l][0.45]{LR}
\psfrag{AFF}[l][l][0.45]{AFF}
\psfrag{Load}[l][l][0.45]{Load}
\psfrag{6}[][][0.45]{6}
\psfrag{12}[][][0.45]{12}
\psfrag{18}[][][0.45]{18}
\psfrag{24}[][][0.45]{24}
\psfrag{30}[][][0.45]{30}
\psfrag{36}[][][0.45]{36}
\psfrag{42}[][][0.45]{42}
\psfrag{48}[][][0.45]{48}
\psfrag{14}[][][0.45]{14}
\psfrag{16}[][][0.45]{16}
\psfrag{20}[][][0.45]{20}
\psfrag{22}[][][0.45]{22}
        \includegraphics[width=1\textwidth]{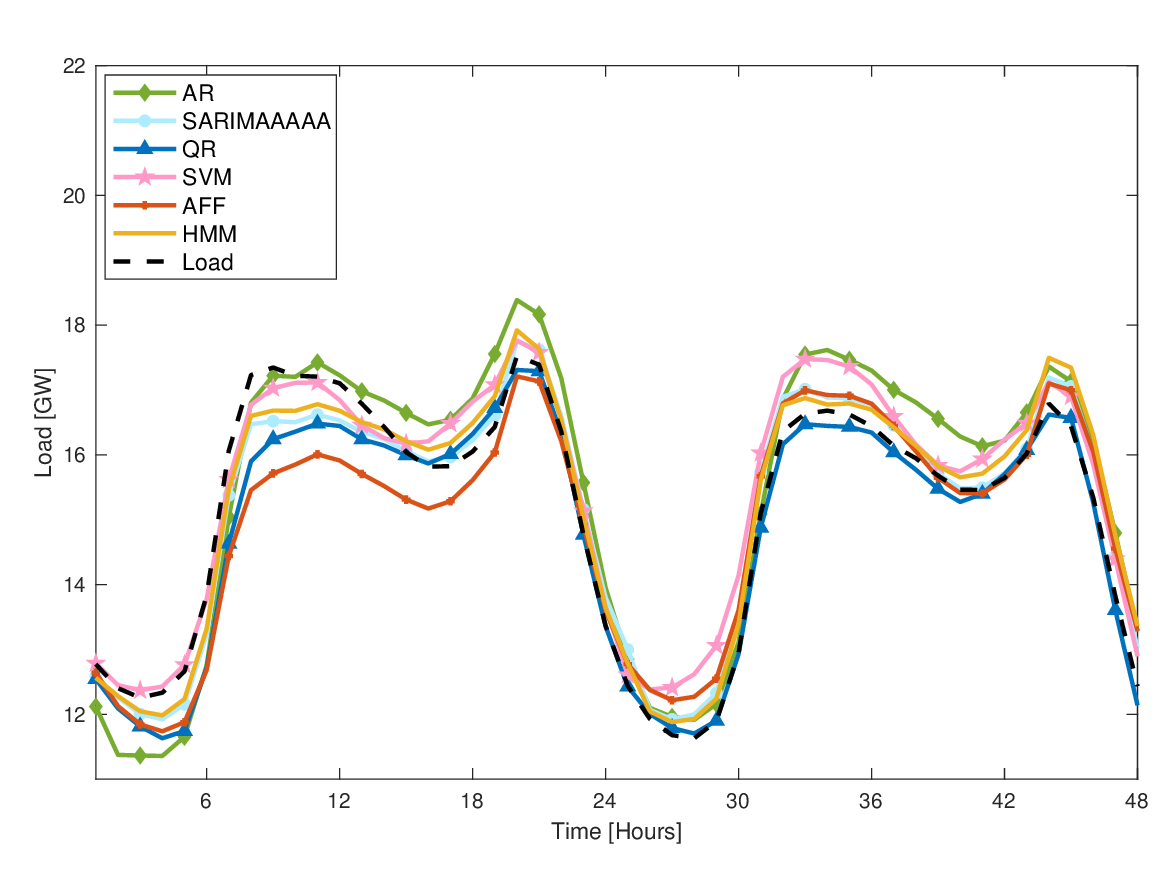}
        \captionsetup{justification=centering}
        \caption{New England.}
    \end{subfigure}\hspace{0.07cm}%
     \begin{subfigure}[t]{0.33\textwidth}
        \centering
               \psfrag{0.9}{{}}
                               \psfrag{steps ahead forecasts}[b][][0.6]{{Prediction horizon}}
        \psfrag{Load [GW]}[b][][0.6]{{Load [GW]}}
        \psfrag{Time [Hours]}[t][][0.6]{{Time [Hours]}}
\psfrag{HMM}[l][l][0.45]{APLF}
\psfrag{SVM}[l][l][0.45]{SVM}
\psfrag{QR}[l][l][0.45]{QR}
\psfrag{SARIMAAAAA}[l][l][0.45]{SARIMA}
\psfrag{AR}[l][l][0.45]{AR}
\psfrag{LR}[l][l][0.45]{LR}
\psfrag{AFF}[l][l][0.45]{AFF}
\psfrag{Load}[l][l][0.45]{Load}
\psfrag{6}[][][0.45]{6}
\psfrag{12}[][][0.45]{12}
\psfrag{18}[][][0.45]{18}
\psfrag{24}[][][0.45]{24}
\psfrag{30}[][][0.45]{30}
\psfrag{36}[][][0.45]{36}
\psfrag{42}[][][0.45]{42}
\psfrag{48}[][][0.45]{48}
\psfrag{1.5}[][][0.45]{1.5}
\psfrag{1.8}[][][0.45]{1.8}
\psfrag{2.1}[][][0.45]{2.1}
\psfrag{2.4}[][][0.45]{2.4}
\psfrag{2.7}[][][0.45]{2.7}
\psfrag{3}[][][0.45]{3}
          \includegraphics[width=1\textwidth]{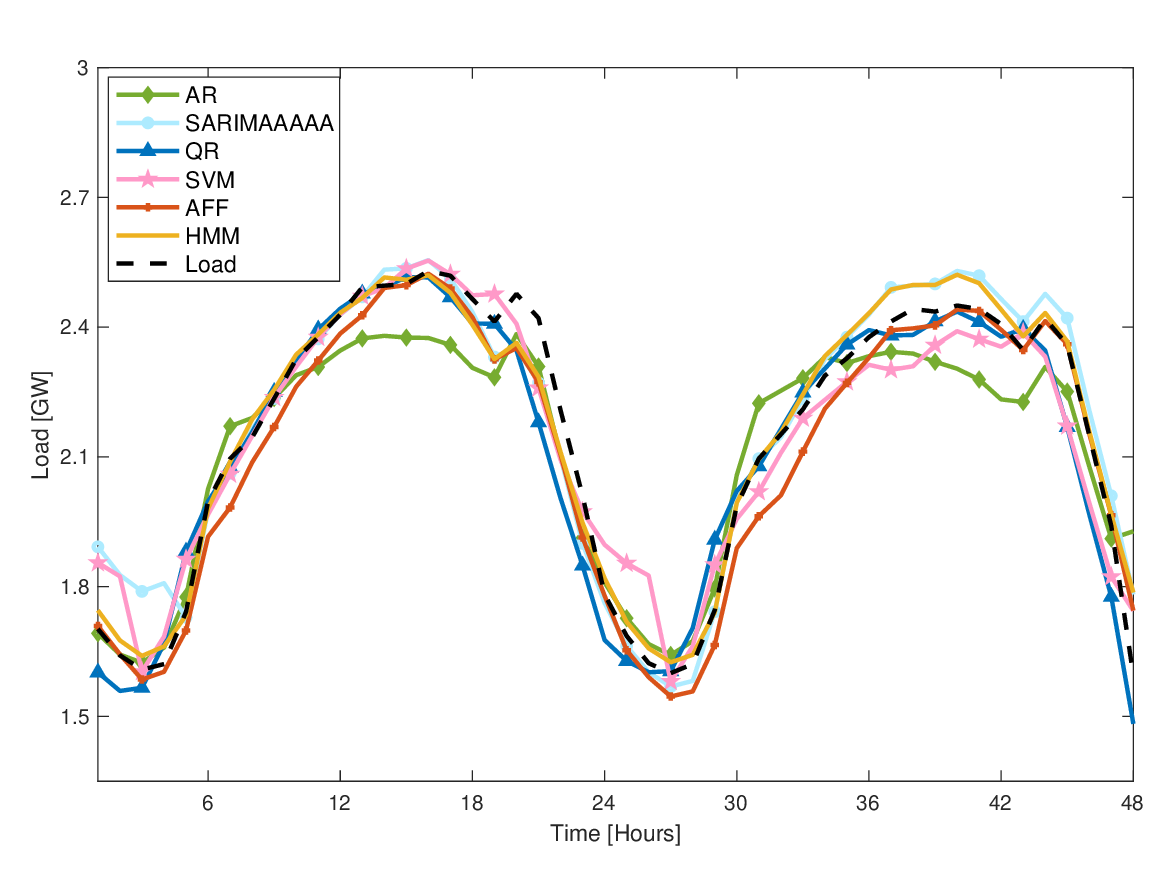}
         \captionsetup{justification=centering}
        \caption{Dayton.}
    \end{subfigure}\hspace{-0.07cm}
    \begin{subfigure}[t]{0.33\textwidth}
        \centering
               \psfrag{9}{{}}
                                              \psfrag{steps ahead forecasts}[b][][0.6]{{Prediction horizon}}
      \psfrag{Load [GW]}[b][][0.6]{{Load [kW]}}
        \psfrag{Time [Hours]}[t][][0.6]{{Time [Hours]}}
\psfrag{HMM}[l][l][0.45]{APLF}
\psfrag{SVM}[l][l][0.45]{SVM}
\psfrag{QR}[l][l][0.45]{QR}
\psfrag{SARIMAAAAA}[l][l][0.45]{SARIMA}
\psfrag{AR}[l][l][0.45]{AR}
\psfrag{LR}[l][l][0.45]{LR}
\psfrag{AFF}[l][l][0.45]{AFF}
\psfrag{Load}[l][l][0.45]{Load}
\psfrag{6}[][][0.45]{6}
\psfrag{12}[][][0.45]{12}
\psfrag{18}[][][0.45]{18}
\psfrag{24}[][][0.45]{24}
\psfrag{30}[][][0.45]{30}
\psfrag{36}[][][0.45]{36}
\psfrag{42}[][][0.45]{42}
\psfrag{48}[][][0.45]{48}
\psfrag{0.2}[][][0.45]{0.2}
\psfrag{0.3}[][][0.45]{0.3}
\psfrag{0.4}[][][0.45]{0.4}
\psfrag{0.5}[][][0.45]{0.5}
\psfrag{0.6}[][][0.45]{0.6}
\psfrag{0.7}[][][0.45]{0.7}
        \includegraphics[width=1\textwidth]{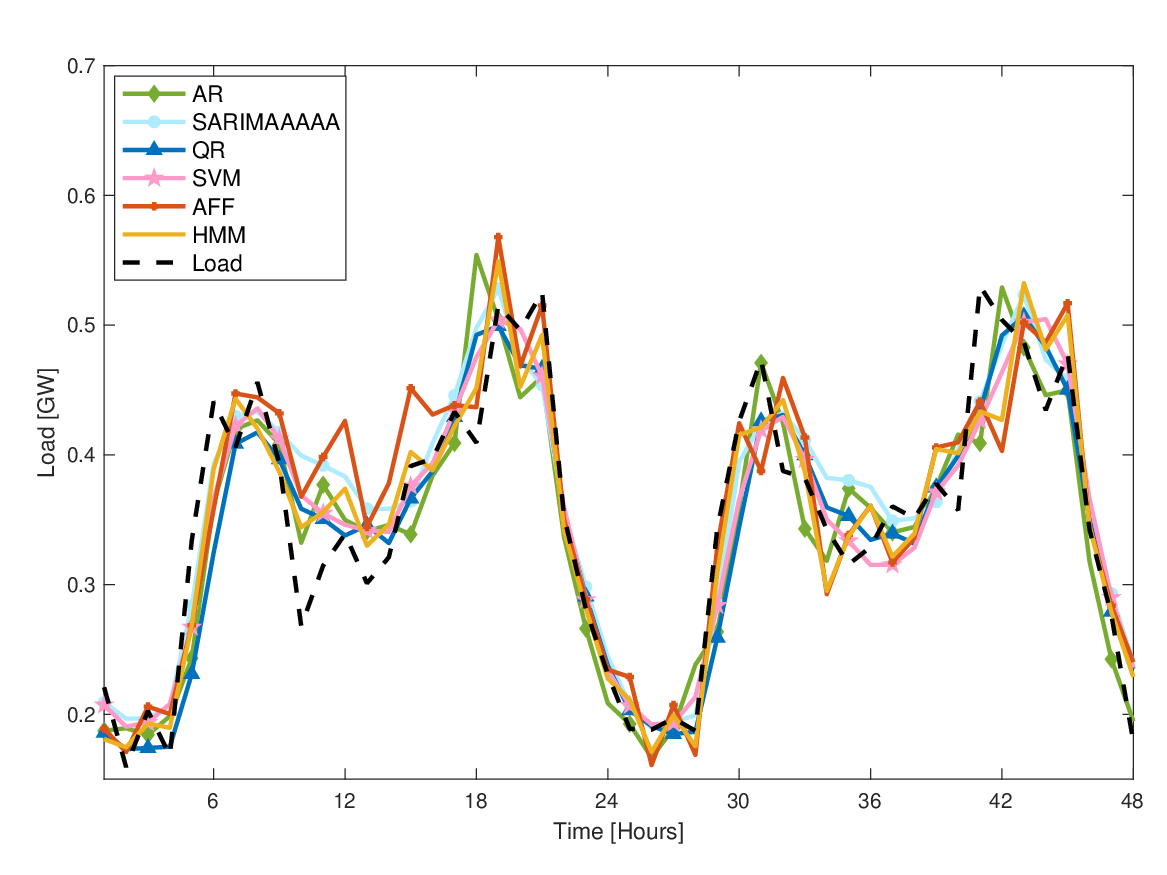}
         \captionsetup{justification=centering}
        \caption{$100$ buildings.}
    \end{subfigure}
    \caption{Load demand presents different consumption patterns and variability in the three regions with different sizes.}
        \label{predictions}
\end{figure*}

We group the seven datasets by size of the region: large, medium, and small. Two datasets belong to large-size regions: load demand in Belgium from 2017-2019 made available by Elia group, and load demand in New England from 2003-2014 made available by ISO-NE organization. Three datasets belong to medium-size regions: Global Energy Forecasting Competition 2012 (GEFCom2012) dataset from 2004-2007 \cite{dataGEF2012}, Global Energy Forecasting Competition 2014 (GEFCom2014) dataset from 2005-2011 \cite{dataGEF2014}, and load demand in Dayton from 2004-2016 made available by PJM interconnection. Finally, two datasets belong to small-size regions that correspond with load demand for $400$ and $100$ buildings in New South Wales from 2013 and are made available by the Australian Government as part of the project Smart Grid Smart Cities. 

\begin{figure*}
    \centering
    \begin{subfigure}[t]{0.33\textwidth}
        \centering
                \psfrag{0.9}{{}}
        \psfrag{y}[b][][0.6]{{CDF}}
        \psfrag{x}[t][][0.6]{{Error [GW]}}
\psfrag{HMM}[l][l][0.45]{APLF}
\psfrag{SVM}[l][l][0.45]{SVM}
\psfrag{QR}[l][l][0.45]{QR}
\psfrag{SARIMAAAAA}[l][l][0.45]{SARIMA}
\psfrag{AR}[l][l][0.45]{AR}
\psfrag{LR}[l][l][0.45]{LR}
\psfrag{AFF}[l][l][0.45]{AFF}
           \psfrag{0.2}[][][0.45]{0.2}
                \psfrag{0.4}[][][0.45]{0.4}
                \psfrag{0.6}[][][0.45]{0.6}
                \psfrag{0}[][][0.45]{0}
                \psfrag{0.8}[][][0.45]{0.8}
                \psfrag{1}[][][0.45]{1}
                \psfrag{0.5}[][][0.45]{0.5}
                 \psfrag{1.5}[][][0.45]{1.5}
 \psfrag{2}[][][0.45]{2}
        \includegraphics[width=1\textwidth]{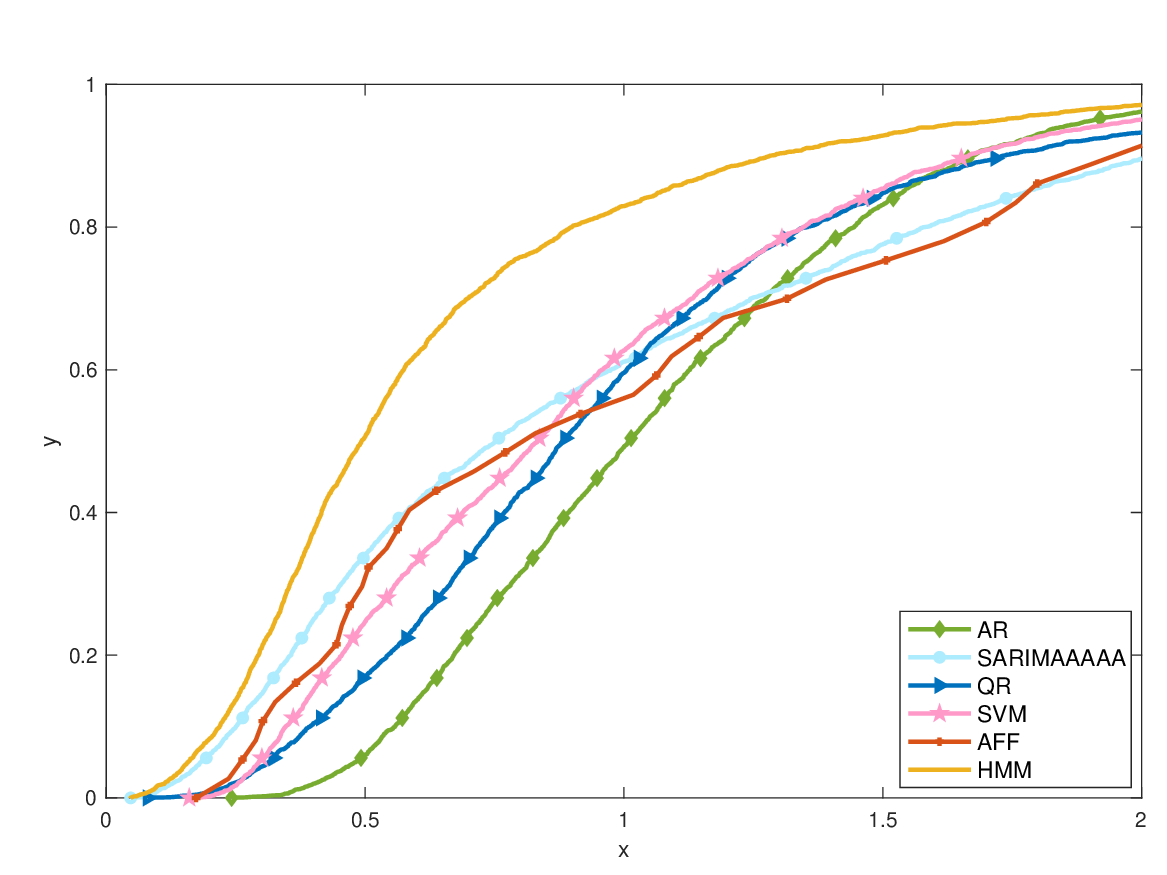}
        \captionsetup{justification=centering}
        \caption{New England.}
    \end{subfigure}\hspace{0.07cm}%
     \begin{subfigure}[t]{0.33\textwidth}
        \centering
               \psfrag{0.9}{{}}
        \psfrag{y}[b][][0.6]{{CDF}}
        \psfrag{x}[t][][0.6]{{Error [GW]}}
\psfrag{HMM}[l][l][0.45]{APLF}
\psfrag{SVM}[l][l][0.45]{SVM}
\psfrag{QR}[l][l][0.45]{QR}
\psfrag{SARIMAAAAA}[l][l][0.45]{SARIMA}
\psfrag{AR}[l][l][0.45]{AR}
\psfrag{LR}[l][l][0.45]{LR}
\psfrag{AFF}[l][l][0.45]{AFF}
           \psfrag{0.2}[][][0.45]{0.2}
                \psfrag{0.4}[][][0.45]{0.4}
                \psfrag{0.6}[][][0.45]{0.6}
                \psfrag{0}[][][0.45]{0}
                \psfrag{0.8}[][][0.45]{0.8}
                \psfrag{1}[][][0.45]{1}
                \psfrag{0.02}[][][0.45]{0.02}
                 \psfrag{0.09}[][][0.45]{0.09}
 \psfrag{0.16}[][][0.45]{0.16}
  \psfrag{0.23}[][][0.45]{0.23}
   \psfrag{0.3}[][][0.45]{0.3}
          \includegraphics[width=1\textwidth]{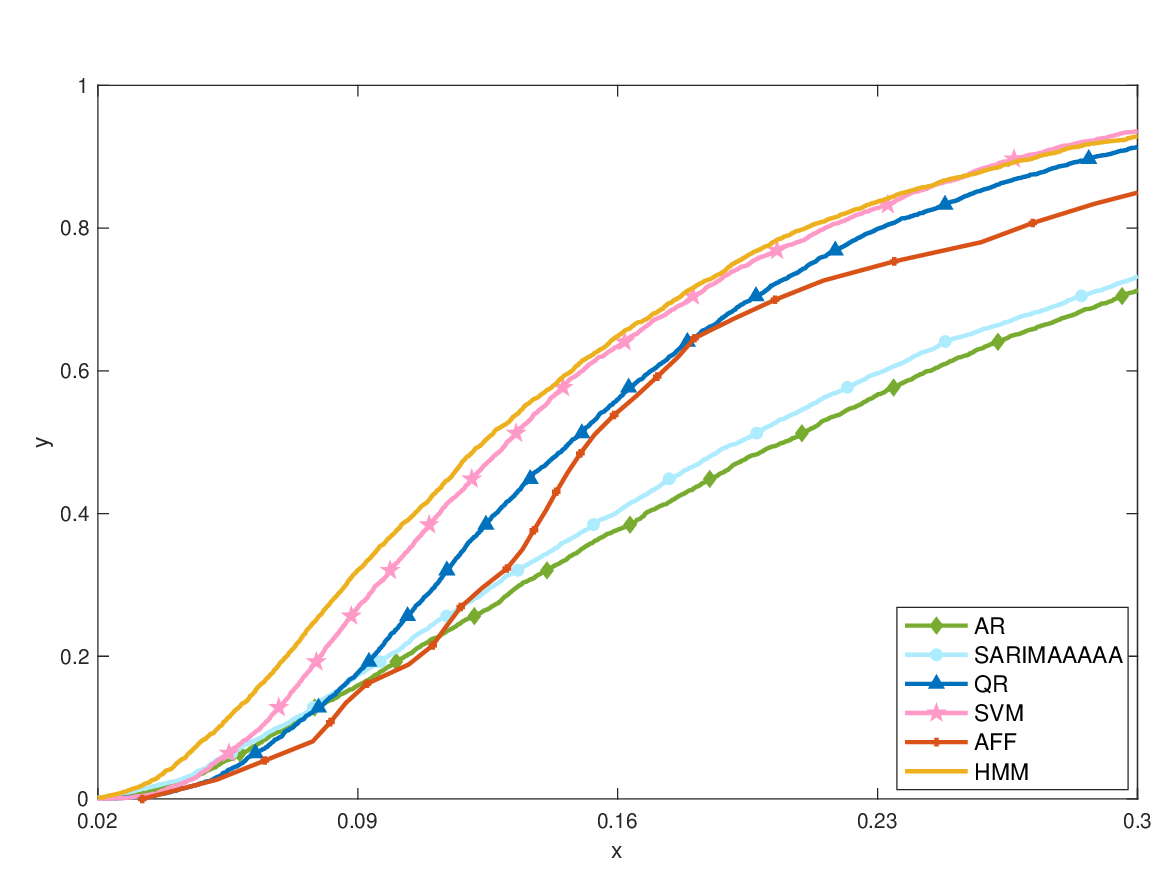}
         \captionsetup{justification=centering}
        \caption{Dayton.}
    \end{subfigure}\hspace{-0.07cm}
    \begin{subfigure}[t]{0.33\textwidth}
        \centering
               \psfrag{9}{{}}
        \psfrag{y}[b][][0.6]{{CDF}}
        \psfrag{x}[t][][0.6]{{Error [kW]}}
\psfrag{HMM}[l][l][0.45]{APLF}
\psfrag{SVM}[l][l][0.45]{SVM}
\psfrag{QR}[l][l][0.45]{QR}
\psfrag{SARIMAAAAA}[l][l][0.45]{SARIMA}
\psfrag{AR}[l][l][0.45]{AR}
\psfrag{LR}[l][l][0.45]{LR}
\psfrag{AFF}[l][l][0.45]{AFF}
           \psfrag{0.2}[][][0.45]{0.2}
                \psfrag{0.4}[][][0.45]{0.4}
                \psfrag{0.6}[][][0.45]{0.6}
                \psfrag{0}[][][0.45]{0}
                \psfrag{0.8}[][][0.45]{0.8}
                \psfrag{1}[][][0.45]{1}
                \psfrag{0.02}[][][0.45]{0.02}
                 \psfrag{0.04}[][][0.45]{0.04}
 \psfrag{0.06}[][][0.45]{0.06}
  \psfrag{0.08}[][][0.45]{0.08}
   \psfrag{0.1}[][][0.45]{0.1}
        \includegraphics[width=1\textwidth]{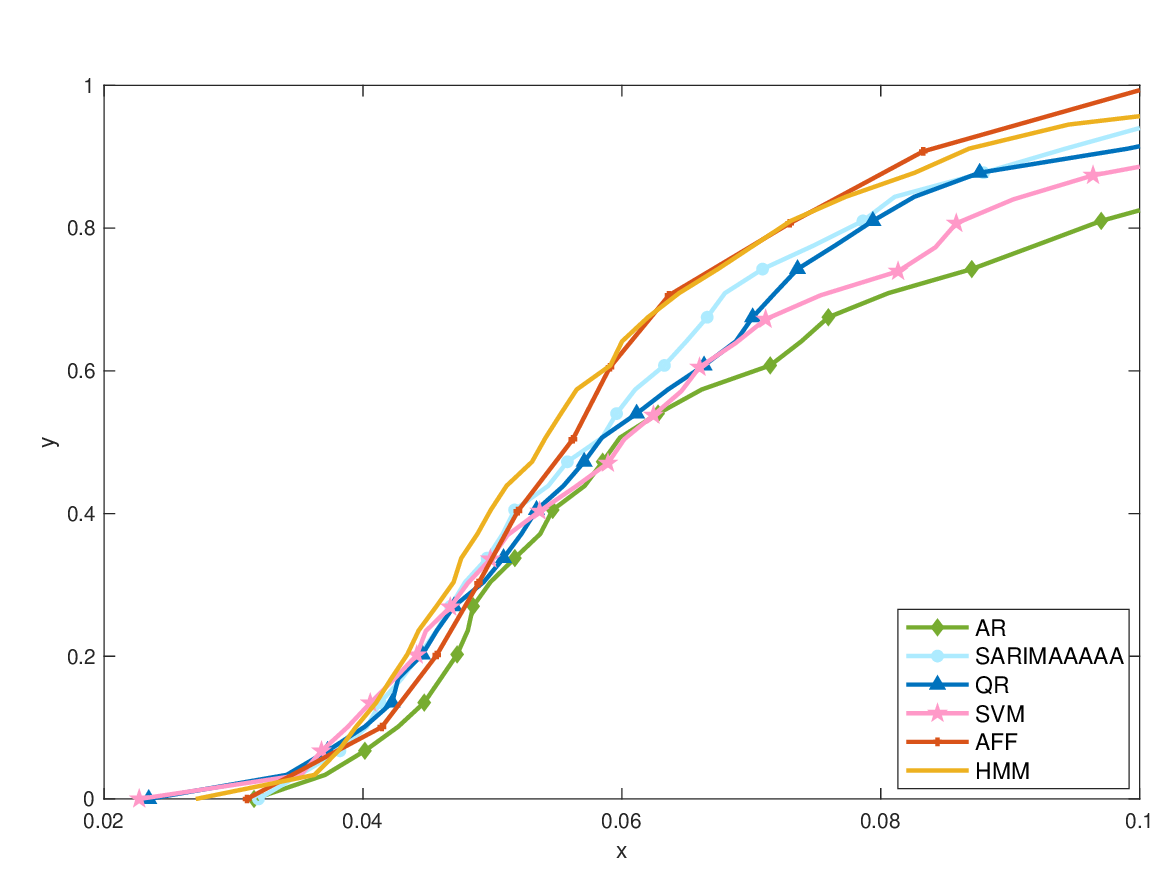}
         \captionsetup{justification=centering}
        \caption{$100$ buildings.}
    \end{subfigure}
    \caption{CDFs of prediction errors provide detailed performance comparison between \ac{APLF} method and conventional techniques in three regions with different sizes.}        \label{ecdf}
\end{figure*}

In the numerical results, training for offline learning algorithms is done using training sizes depending on the length of datasets. Two years of data are used for training in New England and GEFCom2012 datasets; one year of data are used for training in Belgium, GEFCom2014, and Dayton datasets; and $3/4$ of a year of data are used for training in $400$ and $100$ buildings datasets. Prediction for all algorithms is done using the rest of the data as follows. At $11$ a.m. of each day, all forecasting methods obtain future loads for a prediction horizon of $L = 24$ hours hence, every vector of load forecasts is formed by forecasts obtained from $1$ to $L$ hours ahead.

\ac{APLF} results are obtained using the following implementation details. The instance vector composed by past loads and observations is given by $\V{x} = \left[s_t, \V{r}_{t+1}^{\text{T}}, ..., \V{r}_{t+L}^{\text{T}}\right]^{\text{T}}$. The observations vector $\V{r}_t$ contains the temperature $w_t$ at time $t$ and the mean of past temperatures $\bar{w}_{c\left(t\right)}$ at calendar type $c(t)$, i.e., $\V{r}_t = \left[w_t, \bar{w}_{c\left(t\right)}\right]^{\text{T}}$. The observations vector $\V{r}_t$ is represented by the feature vector $u_r\left(\V{r}_t\right)$ for a function $u_r(\cdot)$ that encodes temperature shifts. Specifically, such function determines the vector $u_r\left(\V{r}_t\right) = \left[1, \alpha_1, \alpha_2\right]^{\text{T}}$, where $\alpha_1$ (resp. $\alpha_2$) takes value $1$ if the temperature is above (resp. below) certain thresholds and takes value $0$ otherwise, that is 
\begin{align*}
\alpha_1, \alpha_2 = \left\{ \begin{matrix}1, 0 & \text{if} & w_t - \bar{w}_{c\left(t\right)} & > & W_1 & \text{and} \\
& & w_t > W_2 & \text{or} & w_t < W_3 & \\
0, 1 & \text{if} & w_t - \bar{w}_{c\left(t\right)} & < & - W_1 & \text{and} \\
& & w_t > W_2 & \text{or} & w_t < W_3 & \\
0, 0 & & \text{otherwise} & & && & \\
\end{matrix} \right.
\end{align*}
where we take threshold values $W_1 = 20^{\circ}F$, $W_2 = 80^{\circ}F$, and $W_3 = 20^{\circ}F$ for all datasets. The calendar information $c(t)$ specifies the type of hour: $c(t)$ from $1$ to $24$ indicates the hour of day of weekdays, and $c(t)$ from $25$ to $48$ indicates the hour of day of weekends and holidays, i.e., $c\left(t\right) \in \left\{1, ..., C\right\}$ with $C = 48$. Then, \ac{APLF} method obtains parameters $\B{\eta}_{s, c}, \sigma_{s, c}$ and $\B{\eta}_{r, c}, \sigma_{r, c}$  for $C = 48$ calendar types as given by Algorithm~\ref{alg1} initialized with $\B{\eta}_0 = \V{0}_K$, any $\sigma_0$, $\V{P}_0 = \V{I}_K$, and $\gamma_0 = 0$. Such parameters are updated by taking forgetting factors as $\lambda_s~=~0.2$ and $\lambda_r = 0.7$ for any calendar type and for all datasets.\footnote{Possible numerical instabilities of state matrices $\V{P}$ in Algorithm~\ref{alg1} are addressed by their reinitialization in case their trace becomes larger than $10$ similarly to methods based on recursive least squares \cite{stability}.} Forgetting factors ($\lambda_s, \lambda_r$) and threshold values ($W_1, W_2, W_3$) are the hyper-parameters of APLF method. Hyper-parameters' values can be selected by using various methods such as cross-validation over a grid of possible values. For simplicity, in this paper we select values for hyper-parameters by inspection over one dataset and then we use the same values in all datasets. The numerical results corroborate the robustness of APLF method to the choice of hyper-parameters since we use the same hyper-parameters' values in all numerical results and with significantly different datasets.

\ac{APLF} method is compared with $11$ state-of-the-art techniques based on statistical methods, machine learning, and weighted combination of several forecasts. Three techniques are based on statistical methods: LR \cite{narx}, SARIMA \cite{sarima}, and \ac{QR} \cite{quantilRegression}; six techniques are based on machine learning: \ac{GP} \cite{gp}, \ac{SVM} \cite{svm}, deep residual network (DRN) \cite{deep}, and three versions of AR-NARX method \cite{narx} based on linear regression (AR) \cite{narx}, fixed size least squares \ac{SVM} using the Nystr\"om method (ARNFS) \cite{fs}, and fixed size least squares \ac{SVM} using Random Features (ARRFFS) \cite{rf}; finally, two techniques are based on weighted combination of several forecasts: \ac{SFDA} \cite{hybrid2} and \ac{AFF} \cite{adaptivelr}.

In the first set of numerical results we quantify the prediction error of \ac{APLF} in comparison with the $11$ existing techniques for the $7$ datasets. RMSE and MAPE assessing overall prediction errors are given in Table~\ref{tabla1}. Such table shows that existing techniques such as DRN and \ac{QR} can achieve high accuracies in certain large-size regions using sizeable training datasets (e.g., New England dataset), however such techniques become inaccurate in other datasets such as those corresponding with small-size regions and smaller training datasets (e.g., $100$ buildings dataset). Table~\ref{tabla1} also shows that the online learning method  \ac{AFF} achieves higher accuracies than multiple offline learning algorithms such as LR, ARRFFS, and SFDA. Figures~\ref{predictions} and~\ref{ecdf} provide more detailed comparisons using $5$ representative existing techniques (AR, SARIMA, QR, SVM, and \ac{AFF}) in comparison with proposed APLF in $3$ datasets that correspond with regions of assorted sizes.  Figure~\ref{predictions}  shows two days of load demand and load forecasts in the three regions while Figure~\ref{ecdf} shows the empirical cumulative distribution functions (CDFs) of the absolute value of prediction errors. Table~\ref{tabla1}, and Figures~\ref{predictions} and \ref{ecdf} show that the proposed \ac{APLF} method achieves high accuracies in comparison with existing techniques in every dataset studied. In particular, Figure~\ref{ecdf} shows that high errors occur with low probability for \ac{APLF} method. For instance, in New England dataset, the error of \ac{APLF} method is less than $0.8$~GW with probability $0.8$, while the $5$ other methods reach errors of around $1.3$ GW with such probability. 

\begin{table}
\centering
\caption{Pinball loss and ECE for {APLF} and $2$ state-of-the-art techniques on $7$ datasets.}
\small
\setlength{\tabcolsep}{2pt}
\renewcommand{\arraystretch}{1.5}
\begin{tabular}{|c|ll|ll|ll|}
\hline
\multirow{2}{*}{Region}  &  \multicolumn{2}{c|}{QR} & \multicolumn{2}{c|}{GP} &  \multicolumn{2}{c|}{APLF}\\
& Pinball loss & ECE & Pinball loss & ECE & Pinball loss & ECE \\ \hline
 Belgium & 0.34 [GW] & 0.08 & 0.14 [GW] & 0.19  & \textbf{0.11} [GW] & \textbf{0.07} \\
New Engld. & 0.70 [GW] & \textbf{0.07} & 0.24 [GW] & 0.09 & \textbf{0.22} [GW] & \textbf{0.07} \\ \hline
 GEFCom12 & 1.03 [MW] & \textbf{0.06} & 0.78 [MW] & 0.14 & \textbf{0.77} [MW] & 0.12 \\
GEFCom14 & 0.06 [MW] & 0.60 & \textbf{0.05} [MW] & \textbf{0.15} & 0.06 [MW] & 0.19 \\
Dayton & 0.09 [GW] & 0.06 & \textbf{0.04} [GW] & 0.12 & \textbf{0.04} [GW] & \textbf{0.05} \\ \hline
400 builds.  & 0.02 [kW] & 0.10 & \textbf{0.01} [kW] & \textbf{0.05} & \textbf{0.01} [kW] & 0.07   \\ 
 100 builds. & 0.03 [kW] &0.08 &\textbf{0.01} [kW] & \textbf{0.03} & \textbf{0.01} [kW] & 0.08 \\ \hline
\end{tabular}
\label{tabla2}
\end{table}

   \begin{figure}
\centering
\psfrag{x}[t][][0.6]{Pinball losses [MW]}
\psfrag{y}[b][][0.6]{CDF}
\psfrag{QR}[l][l][0.6]{QR}
\psfrag{GP}[l][l][0.6]{GP}
\psfrag{Benchmarkkkkkkk}[l][l][0.6]{Benchmark}
\psfrag{HMM}[l][l][0.6]{APLF}
\psfrag{0}[][][0.6]{0}
\psfrag{0.2}[][][0.6]{0.2}
\psfrag{0.4}[][][0.6]{0.4}
\psfrag{0.6}[][][0.6]{0.6}
\psfrag{0.8}[][][0.6]{0.8}
\psfrag{1}[][][0.6]{1}
\psfrag{0.04}[][][0.6]{0.04}
\psfrag{0.08}[][][0.6]{0.08}
\psfrag{0.12}[][][0.6]{0.12}
\psfrag{0.16}[][][0.6]{0.16}
\psfrag{0.20}[][][0.6]{0.20}
\includegraphics[width=0.48\textwidth]{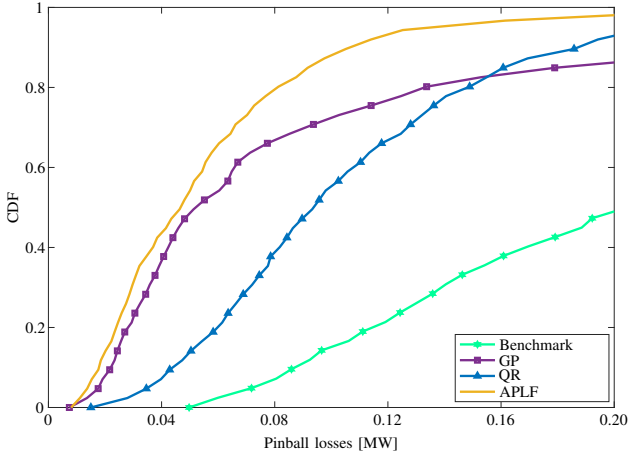}
\caption{CDFs of pinball losses compare the probabilistic performance of \ac{APLF} method with state-of-the-art probabilistic techniques.}
\label{pinbalerror}
\end{figure}

\begin{figure*}
    \centering
         \begin{subfigure}[t]{0.33\textwidth}
        \centering
\psfrag{Probability}[t][][0.6]{Calibration $C(q)$}
        \psfrag{Percentile}[b][][0.6]{Quantile $q$}
                 \psfrag{Perfectly calibratedkkkkkkk}[l][l][0.5]{Perfectly calibrated}
                \psfrag{GP}[l][l][0.5]{GP}
                \psfrag{QR}[l][l][0.5]{QR}
                \psfrag{HMM}[l][l][0.5]{APLF}
                \psfrag{0.2}[][][0.45]{0.2}
                \psfrag{0.4}[][][0.45]{0.4}
                \psfrag{0.6}[][][0.45]{0.6}
                \psfrag{0}[][][0.45]{0}
                \psfrag{0.8}[][][0.45]{0.8}
                \psfrag{1}[][][0.45]{1}
        \includegraphics[width=1\textwidth]{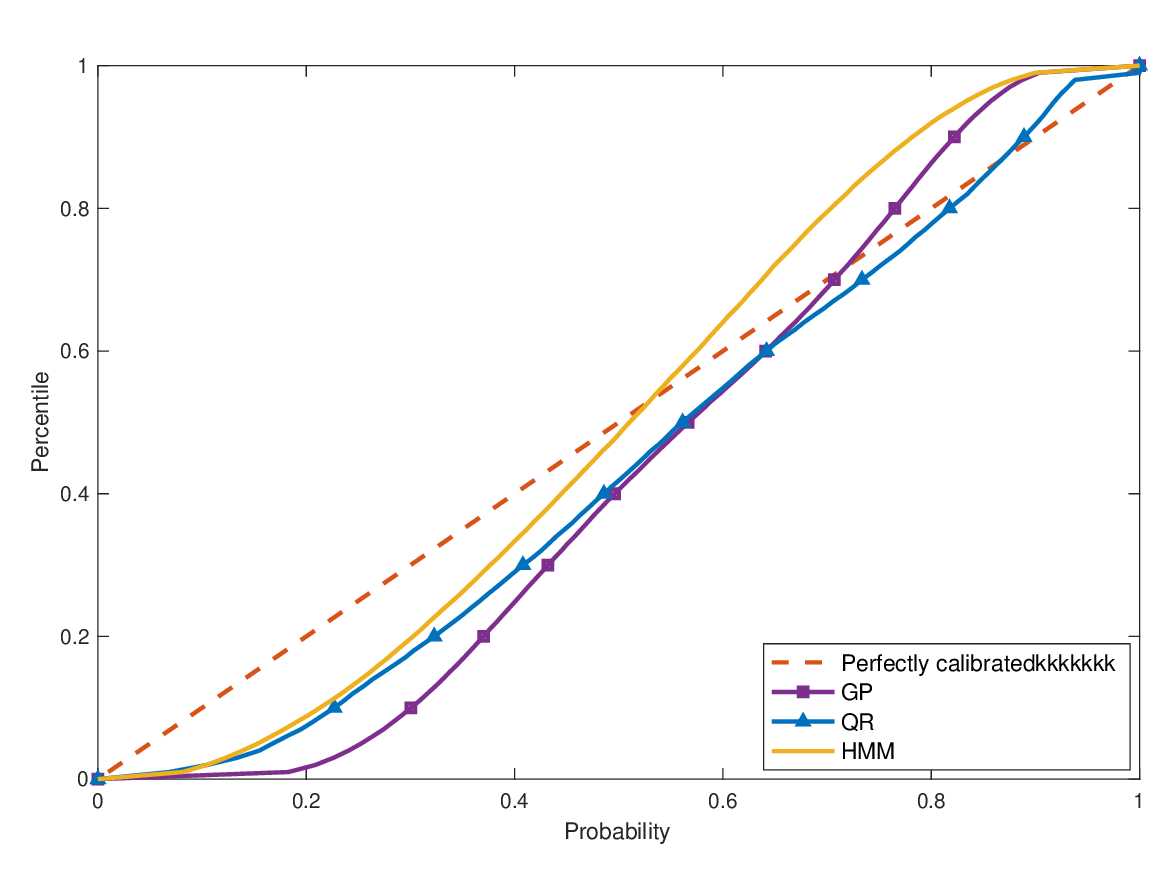}
        \captionsetup{justification=centering}
        \caption{New England.}
    \end{subfigure}\hspace{0.07cm}%
             \begin{subfigure}[t]{0.33\textwidth}
        \centering
      \psfrag{Probability}[t][][0.6]{Calibration $C(q)$}
        \psfrag{Percentile}[b][][0.6]{Quantile $q$}
          \psfrag{Perfectly calibratedkkkkkkk}[l][l][0.5]{Perfectly calibrated}
                \psfrag{GP}[l][l][0.5]{GP}
                \psfrag{QR}[l][l][0.5]{QR}
                \psfrag{HMM}[l][l][0.5]{APLF}
                           \psfrag{0.20}[][][0.45]{0.2}
                           \psfrag{0.2}[][][0.45]{0.2}
                \psfrag{0.40}[][][0.45]{0.4}
                                \psfrag{0.4}[][][0.45]{0.4}
                \psfrag{0.6}[][][0.45]{0.6}
                \psfrag{0}[][][0.45]{0}
                \psfrag{0.80}[][][0.45]{0.8}
                                \psfrag{0.8}[][][0.45]{0.8}
                \psfrag{1}[][][0.45]{1}
        \includegraphics[width=1\textwidth]{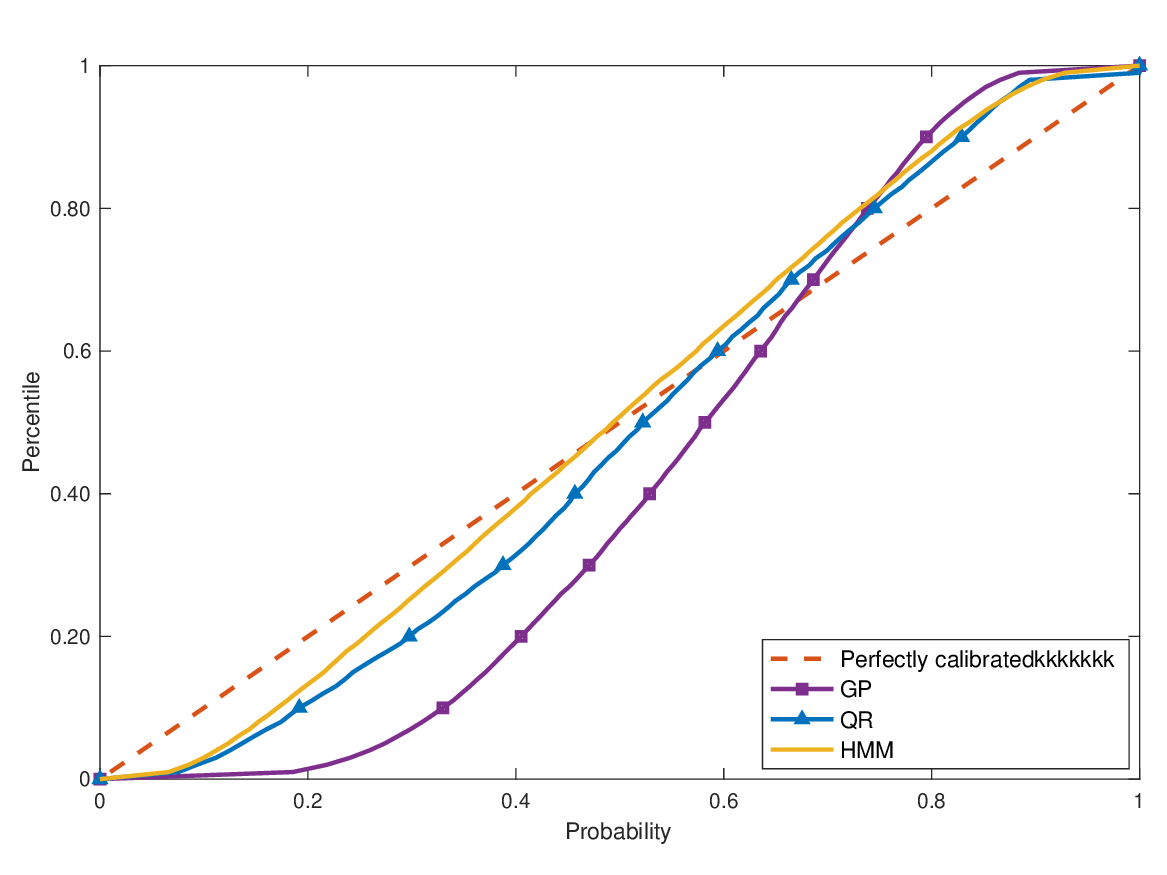}
         \captionsetup{justification=centering}
        \caption{Dayton.}
    \end{subfigure}\hspace{-0.07cm}
     \begin{subfigure}[t]{0.33\textwidth}
        \centering
\psfrag{Probability}[t][][0.6]{Calibration $C(q)$}
        \psfrag{Percentile}[b][][0.6]{Quantile $q$}
           \psfrag{Perfectly calibratedkkkkkkk}[l][l][0.5]{Perfectly calibrated}
                \psfrag{GP}[l][l][0.5]{GP}
                \psfrag{QR}[l][l][0.5]{QR}
                \psfrag{HMM}[l][l][0.5]{APLF}
                           \psfrag{0.2}[][][0.45]{0.2}
                \psfrag{0.4}[][][0.45]{0.4}
                \psfrag{0.6}[][][0.45]{0.6}
                \psfrag{0}[][][0.45]{0}
                \psfrag{0.8}[][][0.45]{0.8}
                \psfrag{1}[][][0.45]{1}
        \includegraphics[width=1\textwidth]{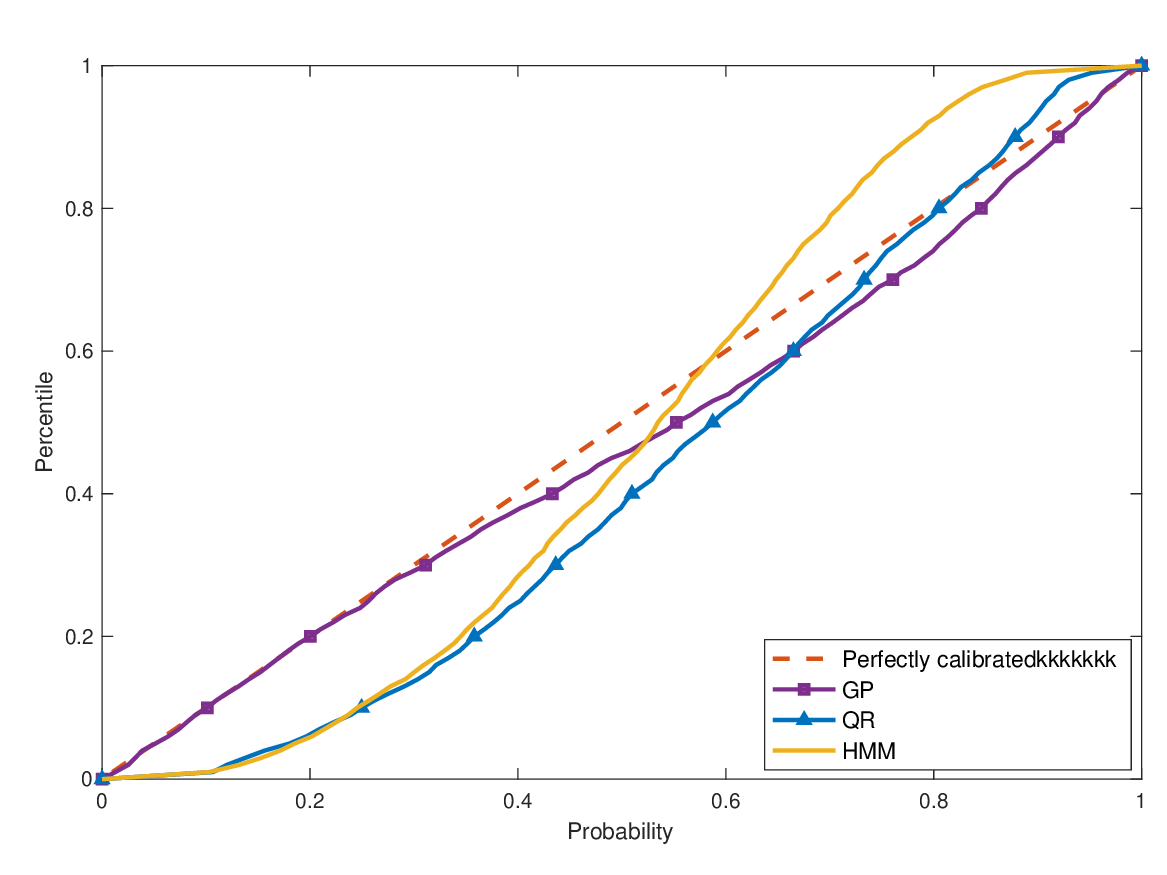}
         \captionsetup{justification=centering}c
        \caption{$100$ buildings.}
    \end{subfigure}\\
       \caption{Calibration plots describe the reliability of probabilistic forecasts of APLF method in comparison with state-of-the-art probabilistic techniques in three regions with different sizes.}
     \label{calib}
    \end{figure*}

 In the second set of numerical results we quantify the probabilistic performance of \ac{APLF} in comparison with \ac{QR} \cite{quantilRegression} and \ac{GP} \cite{gp} and we study the relationship between training size and prediction error. Pinball loss and ECE assessing probabilistic forecasts are given in Table~\ref{tabla2}. Such table shows that APLF achieves high performance in terms of both pinball loss and ECE, while \ac{GP} sometimes achieves poor results in terms of ECE and \ac{QR} achieves poor results in terms of pinball loss. Figures~\ref{pinbalerror} and \ref{calib} provide more detailled quantification of the probabilistic performance of different methods. Figure~\ref{pinbalerror} shows the empirical CDFs of pinball losses of \ac{APLF} method, \ac{QR}, \ac{GP}, and the benchmark for the GEFCom2014 dataset~\cite{dataGEF2014}. These CDFs show that the probability of high pinball losses is significantly lower for \ac{APLF} method. In particular, the CDFs in Figure~\ref{pinbalerror} show that \ac{APLF} and \ac{GP} have a similar median pinball loss of around $0.04$ MW. However, \ac{APLF} has pinball losses less than $0.08$ MW with probability $0.8$, while \ac{GP} reaches pinball losses of $0.16$ MW with such probability. Figure~\ref{calib} shows the correspondence between the calibration $C(q)$ of probabilistic forecasts and the quantile $q$ for the datasets used in Figures~\ref{predictions} and \ref{ecdf}. These calibration plots show that GP and QR tend to obtain forecast quantiles higher than the true quantiles, while APLF obtains more unbiased probabilistic forecasts. In particular, the true load is higher than the 50 quantile forecast load with probability  very near 50 $\%$ for APLF. In addition, Figure~\ref{calib} shows that APLF obtains improved calibrations especially in the lower quantiles.
 
 \begin{figure}
\centering
\psfrag{y}[b][][0.6]{RMSE [MW]}
        \psfrag{x}[t][][0.6]{$t_0$ [Days]}
                \psfrag{HMM}[l][l][0.6]{APLF}
\psfrag{SVM}[l][l][0.6]{SVM}
\psfrag{QR}[l][l][0.6]{QR}
\psfrag{SARIMAAAAAA}[l][l][0.6]{SARIMA}
\psfrag{AR}[l][l][0.6]{AR}
\psfrag{LR}[l][l][0.6]{LR}
\psfrag{AFF}[l][l][0.6]{AFF}
\psfrag{2}[][][0.6]{2}
\psfrag{3.5}[][][0.6]{3.5}
\psfrag{5}[][][0.6]{5}
\psfrag{6.5}[][][0.6]{6.5}
\psfrag{8}[][][0.6]{8}
\psfrag{100}[][][0.6]{100}
\psfrag{200}[][][0.6]{200}
\psfrag{300}[][][0.6]{300}
        \includegraphics[width=0.48\textwidth]{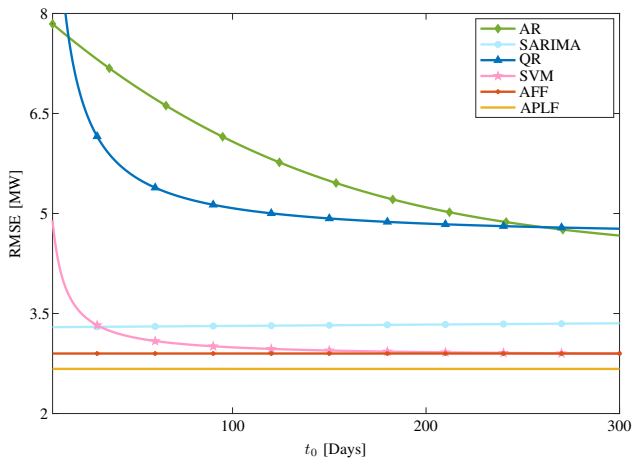}
\caption{The length of the training dataset does not affect the accuracy of online learning algorithms but significantly affects the accuracy of offline learning algorithms.}
\label{training}
\end{figure}
    
    Finally, Figure~\ref{training} shows the RMSE obtained by \ac{APLF} method and the $5$ existing techniques shown in Figures~\ref{predictions} and \ref{ecdf} for different sizes of training sets using GEFCom2012 dataset. These results are obtained computing RMSEs $20$ times for each size of training set. The samples used for training in these numerical results are different at each experiment and testing sets always contain two years of data. As can be observed from Figure~\ref{training}, the accuracy of online learning algorithms does not significantly change with the length of the training dataset, while offline learning algorithms require large training datasets to achieve accurate results.
            
\ac{APLF} method achieves remarkable results both in terms of single-value and probabilistic forecasts, and adapts to different consumption patterns in every region studied even where variability in load demand is more significant. Numerical results confirm that APLF better captures dynamic changes in consumption patterns than existing methods.

\section{Conclusion} \label{concl}

The paper proposes techniques for adaptive probabilistic load forecasting (APLF) that can adapt to changes in consumption patterns and assess load uncertainties. We developed online learning techniques that update model parameters using a simple recursive algorithm, and prediction techniques that obtain probabilistic forecasts using the most recent parameters. In addition, we described the theoretical guarantees and efficient implementation of the online learning and probabilistic prediction steps for \ac{APLF}. The paper also compared the accuracy of the proposed \ac{APLF} with existing techniques in multiple datasets. These datasets represent challenging scenarios with different sizes and different consumption patterns that change over time. The experimental results show the performance improvement of \ac{APLF} method in terms of prediction errors and probabilistic forecasts. As shown in the paper, the proposed method can improve forecasting performance in a wide range of scenarios using efficient and flexible algorithms for adaptive online learning.

\begin{appendices}
\section{Proof Of Theorem \ref{parameters}} \label{app2}

\begin{IEEEproof}
We first prove that for any $i > 0$ the optimal parameters $\B{\eta}_i^*$, $\sigma_i^*$ satisfy
\begin{align}
\label{eta111}
\underset{j = 1}{\overset{i}{\sum}} \lambda^{i-j} \V{u}_{t_j} \V{u}_{t_j}^{\text{T}}\B{\eta}_i^* & = \V{q}_i \\
\label{sigmaestrella}
\gamma_i {\sigma_i^*}^2 & = \underset{j = 1}{\overset{i}{\sum}} \lambda^{{i}-j} s_{t_j}^2 - \V{q}_i^{\text{T}} \B{\eta}_i^*
\end{align}while parameters $\B{\eta}_i$, $\sigma_i$ and matrix $\V{P}_i$ given by recursions \eqref{etarecursion5}-\eqref{eqg11} with $\B{\eta}_0 = \V{0}_K$, any $\sigma_0$, $\V{P}_0 = \V{I}_K$, and $\gamma_0 = 0$ satisfy 
\begin{align}
\label{etai}
\V{P}_i^{-1} \B{\eta}_i & = \V{q}_i \\
\label{sigmai}
\gamma_i {\sigma_i}^2 & = \underset{j = 1}{\overset{i}{\sum}} \lambda^{{i}-j} s_{t_j}^2 - \V{q}_i^{\text{T}} \B{\eta}_i\\
\label{ppestrella}
\V{P}_i^{-1} & =  \lambda^i\V{I}_K +  \underset{j = 1}{\overset{i}{\sum}} \lambda^{i-j} \V{u}_{t_j} \V{u}_{t_j}^{\text{T}}
\end{align}where $\V{q}_i = \underset{j = 1}{\overset{{i}}{\sum}}\lambda^{{i} - j}  s_{t_j} \V{u}_{t_j}$. Then, in the second step of the proof we obtain bound in \eqref{aproxerror} using equations \eqref{eta111}-\eqref{ppestrella}. Finally, we prove that parameters $\B{\eta}_i$, and $\sigma_i$ given by recursions \eqref{etarecursion5}-\eqref{eqg11} with $\B{\eta}_{i_0}$, $\sigma_{i_0}$ given by \eqref{etai0}-\eqref{gammai0} satisfy $\B{\eta}_i = \B{\eta}_i^*$ and $\sigma_i = \sigma_i^*$, for $i \geq i_0$.

Parameters $\B{\eta}_i^*$ and $\sigma_i^*$ satisfy equations \eqref{eta111} and \eqref{sigmaestrella}, respectively, because they maximize the log-likelihood in \eqref{loglike}. The differentiable function $L_i\left(\B{\eta}, \sigma\right)$ is concave since Gaussian functions are log-concave. Then, $L_i(\B{\eta}, \sigma)$ has a maximum achieved by parameters that  result in zero derivatives. Since 
\begin{align*}
L_i\left(\B{\eta}, \sigma\right) & = - \underset{j = 1}{\overset{{i}}{\sum}}\lambda^{{i} - j}\frac{\big(s_{t_j} -  \V{u}_{t_j}^{\text{T}} \B{\eta}\big)^2}{2 \sigma^2} + \lambda^{{i} - j} \log\big(\sigma \sqrt{2 \pi}\big)
\end{align*} 
we have that
\begin{align}
\frac{\partial L_i(\B{\eta}, \sigma)}{\partial \B{\eta}}& = \underset{j = 1}{\overset{{i}}{\sum}}\lambda^{{i} - j} \frac{\V{u}_{t_j} (s_{t_j} - \V{u}_{t_j}^{\text{T}} \B{\eta})}{{\sigma}^2}\nonumber 
\end{align}
that becomes zero for $\B{\eta}_i^*$ given by \eqref{eta111}, and
\begin{align*}
\frac{\partial L_i(\B{\eta}, \sigma)}{\partial \sigma} = \underset{j = 1}{\overset{i}{\sum}}\lambda^{{i}-j} \frac{\big(s_{t_j} -  \V{u}_{t_j}^{\text{T}} \B{\eta}\big)^2}{{\sigma}^3} - \lambda^{{i} - j} \frac{1}{\sigma}
\end{align*}
that becomes zero for $\sigma_i^*$ given by \eqref{sigmaestrella} since $\gamma_i$ given by \eqref{eqg11} equals $\gamma_i = \underset{j = 1}{\overset{i}{\sum}}\lambda^{{i}-j}$.

By induction, we prove that $\B{\eta}_i$ and $\sigma_i$ given by recursions (\ref{etarecursion5}) and (\ref{eqsigmarecursion21}) satisfy equations (\ref{etai}) and (\ref{sigmai}) for $\B{\eta}_0 = \V{0}_K$, any $\sigma_0$, $\V{P}_0 = \V{I}_K$, and $\gamma_0 = 0$. Firstly, we prove it for $i = 1$. From \eqref{eqP1}, we have that
\begin{align}
\label{P1}
\V{P}_{1} &  = \frac{1}{\lambda}\left(\V{I}_K -  \frac{\V{u}_{t_1} \V{u}_{t_1}^{\text{T}}}{\lambda + \V{u}_{t_1}^{\text{T}} \V{u}_{t_1}}\right) = \left(\lambda \V{I}_K + \V{u}_{t_1} \V{u}_{t_1}^{\text{T}}\right)^{-1}
\end{align}
applying the matrix inversion Lemma and using that $\V{P}_0 = \V{I}_K$.
Hence, from \eqref{etarecursion5}, \eqref{eqsigmarecursion21}, and \eqref{eqg11}, we get
\begin{align*}
\V{P}_{1}^{-1} \B{\eta}_1 & = \left(\lambda \V{I}_K + \V{u}_{t_1} \V{u}_{t_1}^{\text{T}}\right) \frac{s_{t_1} \V{u}_{t_1}}{\lambda + \V{u}_{t_1}^{\text{T}} \V{u}_{t_1}}  = s_{t_1} \V{u}_{t_1}\\
\sigma_{1}^2 & = \frac{\lambda s_{t_1}^2 }{\lambda + \V{u}_{t_1}^{\text{T}}\V{u}_{t_1}} = \frac{\lambda s_{t_1}^2 + s_{t_1}^2 \V{u}_{t_1}^{\text{T}} \V{u}_{t_1} - s_{t_1}^2 \V{u}_{t_1}^{\text{T}} \V{u}_{t_1}}{\lambda + \V{u}_{t_1}^{\text{T}} \V{u}_{t_1}} \\
& =  s_{t_1}^2 - s_{t_1} \V{u}_{t_1}^{\text{T}} \frac{s_{t_1}\V{u}_{t_1}}{\lambda + \V{u}_{t_1}^{\text{T}} \V{u}_{t_1}} = s_{t_1}^2 - s_{t_1} \V{u}_{t_1}^{\text{T}} \B{\eta}_1
\end{align*}
since $\B{\eta}_0 = \V{0}_K$, $\V{P}_0 = \V{I}_K$, and $\gamma_0 = 0$.

If \eqref{etai} and \eqref{sigmai} hold for $i - 1$, then for $i$ we have that
\begin{align}
\label{Pi}
\V{P}_{i} & = \left(\lambda{\V{P}_{i-1}}^{-1} + \V{u}_{t_i} \V{u}_{t_i}^{\text{T}}\right)^{-1}
\end{align}
applying the matrix inversion Lemma to equation \eqref{eqP1}. Therefore, using the recursion of $\B{\eta}_i$ in \eqref{etarecursion5}, we have that
\begin{align}
\V{P}_i^{-1} \B{\eta}_{i} =& \lambda{\V{P}_{i-1}}^{-1} \B{\eta}_{i-1} +\frac{ \lambda \V{u}_{t_i}}{\lambda + \V{u}_{t_i}^{\text{T}} \V{P}_{i-1} \V{u}_{t_i}} \left(s_{t_i} - \V{u}_{t_i}^{\text{T}} \B{\eta}_{i-1}\right) \nonumber \\
& + \V{u}_{t_i} \V{u}_{t_i}^{\text{T}}\B{\eta}_{i-1} + \frac{\V{u}_{t_i} \V{u}_{t_i}^{\text{T}} \V{P}_{i-1} \V{u}_{t_i}}{\lambda + \V{u}_{t_i}^{\text{T}} \V{P}_{i-1} \V{u}_{t_i}} \left(s_{t_i} - \V{u}_{t_i}^{\text{T}} \B{\eta}_{i-1}\right) \nonumber \\
=& \lambda \V{q}_{i-1} +  \frac{\lambda \V{u}_{t_i}}{\lambda + \V{u}_{t_i}^{\text{T}}\V{P}_{i-1} \V{u}_{t_i}} \left(s_{t_i} - \V{u}_{t_i}^{\text{T}} \B{\eta}_{i-1}\right) \nonumber \\
\label{indhyp}
& + \V{u}_{t_i} \V{u}_{t_i}^{\text{T}}\B{\eta}_{i-1} + \frac{\V{u}_{t_i} \V{u}_{t_i}^{\text{T}}  \V{P}_{i-1}\V{u}_{t_i}}{\lambda + \V{u}_{t_i}^{\text{T}}\V{P}_{i-1} \V{u}_{t_i}} \left(s_{t_i} - \V{u}_{t_i}^{\text{T}} \B{\eta}_{i-1}\right) \\
= & \lambda \V{q}_{i-1} + \V{u}_{t_i} \left(s_{t_i} - \V{u}_{t_i}^{\text{T}} \B{\eta}_{i-1}\right) + \V{u}_{t_i} \V{u}_{t_i}^{\text{T}}\B{\eta}_{i-1} \nonumber \\
= & \underset{j = 1}{\overset{i-1}{\sum}} \lambda^{i-j} s_{t_j} \V{u}_{t_j} + s_{t_i} \V{u}_{t_i} = \V{q}_i \nonumber
\end{align}where the equality \eqref{indhyp} is obtained by using the induction hypothesis.
Using the recursion of $\sigma_i$ in \eqref{eqsigmarecursion21}, we have that


\begin{align}
\gamma_i {\sigma_i}^2 = & \left(\gamma_i - 1\right) {\sigma_{i-1}}^2 + \left(s_{t_i} - \V{u}_{t_i}^{\text{T}} \B{\eta}_{i-1}\right)\nonumber \\
& \cdot \left(s_{t_i} - \frac{ s_{t_i}\V{u}_{t_i}^{\text{T}} \V{P}_{i-1} \V{u}_{t_i}}{\lambda + \V{u}_{t_i}^{\text{T}} \V{P}_{i-1} \V{u}_{t_i}} - \frac{ \lambda \V{u}_{t_i}^{\text{T}}  \B{\eta}_{i-1}}{\lambda + \V{u}_{t_i}^{\text{T}} \V{P}_{i-1} \V{u}_{t_i}}\right) \nonumber \\
\label{anagproc}
= & \underset{j = 1}{\overset{i-1}{\sum}} \lambda^{{i}-j} s_{t_j}^2 - \lambda \V{q}_{i-1}^{\text{T}} \B{\eta}_{i-1} + \left(s_{t_i} - \V{u}_{t_i}^{\text{T}} \B{\eta}_{i-1}\right) \\
 \label{indhyp2}
&\cdot \left(s_{t_i} - \frac{s_{t_i} \V{u}_{t_i}^{\text{T}} \V{P}_{i-1} \V{u}_{t_i}}{\lambda + \V{u}_{t_i}^{\text{T}} \V{P}_{i-1} \V{u}_{t_i}}  -  \frac{\lambda \V{q}_{i-1}^{\text{T}} \V{P}_{i-1} \V{u}_{t_i}}{\lambda + \V{u}_{t_i}^{\text{T}} \V{P}_{i-1} \V{u}_{t_i}} \right)\\
 = &\underset{j = 1}{\overset{i-1}{\sum}} \lambda^{{i}-j} s_{t_j}^2 + s_{t_i}^2 -  \left(\underset{j = 1}{\overset{i-1}{\sum}} \lambda^{i-j} s_{t_j} \V{u}_{t_j}^{\text{T}} + s_{t_i} \V{u}_{t_i}^{\text{T}}\right) \nonumber \\
 \label{finsigma}
 & \cdot \left(\B{\eta}_{i-1}+   \frac{\V{P}_{i-1} \V{u}_{t_i}}{\lambda + \V{u}_{t_i}^{\text{T}} \V{P}_{i-1} \V{u}_{t_i}} \left(s_{t_i} - \V{u}_{t_i}^{\text{T}} \B{\eta}_{i-1}\right) \right)
 \end{align}
where the equality \eqref{indhyp2} is obtained by using the induction hypothesis. Then, we obtain \eqref{sigmai} from \eqref{finsigma} by using the recursion for $\B{\eta}_i$ in \eqref{etarecursion5}.

In addition, by induction we prove that $\V{P}_i$ given by recursion \eqref{eqP1} satisfies the equation (\ref{ppestrella}). The case $i = 1$ is proved in the equation \eqref{P1}. If \eqref{ppestrella} holds for $i-1$, then for $i$ by using~\eqref{Pi} and the induction hypothesis we have that
 \begin{align*}
 \V{P}_{i} & =  \bigg(\lambda\Big(\lambda^{i-1} \V{I}_K + \underset{j = 1}{\overset{i-1}{\sum}} \lambda^{i-1-j} \V{u}_{t_j} \V{u}_{t_j}^{\text{T}}\Big) + \V{u}_{t_i} \V{u}_{t_i}^{\text{T}}\bigg)^{-1}
\end{align*}
that directly leads to (\ref{ppestrella}).

To obtain the bound in \eqref{aproxerror}, we first use the definition of $L_i$ and equations \eqref{sigmaestrella} and \eqref{sigmai} to obtain
\begin{align}
L_i & \left( \B{\eta}_i^*, \sigma_i^*\right) - L_i\left(\B{\eta}_i, \sigma_i\right) = \frac{ \gamma_i }{2} \log\left(\frac{\sigma_i^2}{{\sigma_i^{*}}^2}\right) \nonumber \\
& = \frac{ \gamma_i }{2} \log\left(1 + \frac{\sigma_i^2 - {\sigma_i^{*}}^2}{{\sigma_i^{*}}^2}\right) \leq \frac{ \gamma_i }{2} \frac{\Big{|}\sigma_i^2 - {\sigma_i^{*}}^2\Big{|} }{{\sigma_i^{*}}^2} \nonumber\\
\label{ineqL}
& = \frac{ 1 }{2} \left|\frac{\V{q}_i^{\text{T}} \left(\B{\eta}_i^* - \B{\eta}_i\right)}{{\sigma_i^*}^2}\right|.
\end{align}
We then use the following inequalities
\begin{align}
\label{ineqeta}
 \left|\V{q}_i^{\text{T}} \left(\B{\eta}_i^* - \B{\eta}_i\right)\right| & \leq \lambda^i  \left\|{\B{\eta}_i^*}\right\| \left\|\B{\eta}_i\right\|\\
\label{ineqsigma}
\frac{1}{{\sigma_i^*}^2} & \leq 2 \pi M^2
\end{align}where the inequality \eqref{ineqeta} is obtained using equations \eqref{eta111} and \eqref{etai} because
\begin{align}
\big|&\V{q}_i^{\text{T}} \left(\B{\eta}_i^* - \B{\eta}_i\right)\big| = \Bigg|\V{q}_i^{\text{T}} \Bigg(\bigg(\underset{j = 1}{\overset{i}{\sum}} \lambda^{i-j} \V{u}_{t_j} \V{u}_{t_j}^{\text{T}}\bigg)^{-1}-\V{P}_i\Bigg)\V{q}_i\Bigg| \nonumber\\
& = \Bigg|\V{q}_i^{\text{T}} \bigg(\underset{j = 1}{\overset{i}{\sum}} \lambda^{i-j} \V{u}_{t_j} \V{u}_{t_j}^{\text{T}}\bigg)^{-1}\bigg(\V{P}_i^{-1} -  \underset{j = 1}{\overset{i}{\sum}} \lambda^{i-j} \V{u}_{t_j} \V{u}_{t_j}^{\text{T}}\bigg) \V{P}_i \V{q}_i \Bigg|\nonumber\\
& = \Bigg|\lambda^i \V{q}_i^{\text{T}} \bigg(\underset{j = 1}{\overset{i}{\sum}} \lambda^{i-j} \V{u}_{t_j} \V{u}_{t_j}^{\text{T}}\bigg)^{-1} \V{P}_i \V{q}_i\Bigg| = \lambda^i\left|{\B{\eta}_i^*}^{\text{T}} \B{\eta}_i\right|\nonumber
\end{align}
and the inequality \eqref{ineqsigma} is obtained due to the fact that
\begin{align} 
\label{sigeq}
\log \left(\frac{1}{\sigma_i^*}\right) - \log \sqrt{2 \pi} \leq \log M \nonumber \Rightarrow \left|\frac{1}{\sigma_i^*}\right| \leq \sqrt{2 \pi} M 
\end{align}
because
\begin{equation*}
L_i\left(\B{\eta}_i^*, \sigma_i^*\right) = \gamma_i( - \log \sigma_i^* - \log \sqrt{2\pi})
\end{equation*}
and
\begin{equation*}
\hspace{-2.1cm} L_i\left(\B{\eta}_i^*, \sigma_i^*\right) \leq \gamma_i \log M
\end{equation*}
 since $N(s_{t_j}; \V{u}_{t_j}^{\text{T}} \B{\eta}_i^*, \sigma_i^*) \leq M$ for any $j \leq i \leq n$. 
 
Substituting inequalities \eqref{ineqeta} and \eqref{ineqsigma} in \eqref{ineqL}, we have that
\begin{align*}
L_i \left( \B{\eta}_i^*, \sigma_i^*\right) - L_i\left(\B{\eta}_i, \sigma_i\right)\leq & \pi M^2 \lambda^i  \left\|{\B{\eta}_i^*}\right\| \left\|\B{\eta}_i\right\|
\end{align*}
that leads to bound in \eqref{aproxerror} using the definition of $\B{\eta}_i$ given by \eqref{etai} and the following inequalities
\begin{align*}
\Bigg\|\V{P}_i \underset{j = 1}{\overset{i}{\sum}} \lambda^{i-j} \V{u}_{t_j} \V{u}_{t_j}^{\text{T}} \B{\eta}_i^*\Bigg\| & \leq \Bigg\|\V{P}_i \underset{j = 1}{\overset{i}{\sum}} \lambda^{i-j} \V{u}_{t_j} \V{u}_{t_j}^{\text{T}} \Bigg\| \left\|\B{\eta}_i^*\right\| \leq \left\|\B{\eta}_i^*\right\|
\end{align*}
where the last inequality is obtained because for any $i$ such that the matrix \eqref{nonsmatrix} is not singular, we have that
\begin{align*} 
& \bigg( \underset{j = 1}{\overset{i}{\sum}} \lambda^{i-j} \V{u}_{t_j} \V{u}_{t_j}^{\text{T}} \bigg)^{-1} \bigg(\underset{j = 1}{\overset{i}{\sum}} \lambda^{i-j} \V{u}_{t_j} \V{u}_{t_j}^{\text{T}} + \lambda^i \V{I}_K\bigg) \succeq \V{I}_K
\end{align*}
which implies
\begin{align*}
& \bigg(\underset{j = 1}{\overset{i}{\sum}} \lambda^{i-j} \V{u}_{t_j} \V{u}_{t_j}^{\text{T}} + \lambda^i \V{I}_K\bigg)^{-1} \underset{j = 1}{\overset{i}{\sum}} \lambda^{i-j} \V{u}_{t_j} \V{u}_{t_j}^{\text{T}} \\
& = \V{P}_i \underset{j = 1}{\overset{i}{\sum}} \lambda^{i-j} \V{u}_{t_j} \V{u}_{t_j}^{\text{T}} \preceq \V{I}_K.
\end{align*}

Now, we proof by induction that for any $i \geq i_0$ parameters $\B{\eta}_i$ and $\sigma_i$ given by recursions \eqref{etarecursion5}-\eqref{eqg11} with $\B{\eta}_{i_0}$, $\sigma_{i_0}$, $\V{P}_{i_0}$, and $\gamma_{i_0}$ given by \eqref{etai0}-\eqref{gammai0} satisfy $\B{\eta}_i = \B{\eta}_i^*$, $\sigma_i = \sigma_i^*$, and $\V{P}_{i} = (\underset{j = 1}{\overset{i}{\sum}} \lambda^{i-j} \V{u}_{t_j} \V{u}_{t_j}^{\text{T}})^{-1}$. Firstly, for $i = i_0$ the assertions are obtained directly from \eqref{etai0}-\eqref{gammai0} since $\V{H}_{i_0}$ is non-singular and $\B{\eta}_i^*$ and $\sigma_i^*$ satisfy \ref{eta111} and \ref{sigmaestrella}, respectively.

If $\B{\eta}_{i-1} = \B{\eta}_{i-1}^*$ and $\sigma_{i-1} = \sigma_{i-1}^*$ hold, then for $i$ we have that 
\begin{align}
\label{eqP12}
\V{P}_{i} = (\lambda \V{P}_{i-1}^{-1} + \V{u}_{t_{i}}\V{u}_{t_{i}}^{\text{T}})^{-1} = \bigg(\underset{j = 1}{\overset{i}{\sum}} \lambda^{i-j} \V{u}_{t_j} \V{u}_{t_j}^{\text{T}}\bigg)^{-1}
\end{align}
applying the matrix inversion Lemma to equation \eqref{eqP1}. From \eqref{etarecursion5}, we get
\begin{align*}
\B{\eta}_{i} = & \V{P}_{i-1} \V{q}_{i-1} + \frac{\V{P}_{i-1} \V{u}_{t_{i}}}{\lambda + \V{u}_{t_{i}}^{\text{T}} \V{P}_{i-1} \V{u}_{t_{i}}} \big(s_{t_{i}} - \V{u}_{t_{i}}^{\text{T}}\V{P}_{i-1} \V{q}_{i-1} \big)\\
= &  \frac{\V{P}_{i-1}}{\lambda + \V{u}_{t_{i}}^{\text{T}} \V{P}_{i-1} \V{u}_{t_{i}}} \V{q}_i = \V{P}_{i} \V{q}_i
\end{align*}
by replacing the induction hypothesis and using \eqref{eqP12} together with the matrix inversion Lemma. 
Then, the result for $\sigma_i = \sigma_i^*$ can be obtained analogously to the steps in \eqref{anagproc}-\eqref{finsigma}.
\end{IEEEproof}

\section{Proof of Theorem \ref{teoremaRecursion}} \label{ap3}

The proof uses the following Lemma.
\begin{lema}\label{lema1}
Let $N\left( x; a, b\right)$, $N\left( y;  \alpha x, \beta\right)$ be two Gaussian density functions, then
\begin{align*}
& N\left( x; a, b\right) N\left( y;  \alpha x, \beta\right) = \\
& N\left( x; \frac{a \beta^2 + \alpha y b^2}{\beta^2 + \alpha^2 b^2}, \sqrt{\frac{b^2 \beta^2}{\beta^2 + \alpha^2 b^2}}\right) N\left( y; a \alpha, \sqrt{\beta^2 + \alpha^2 b^2}\right).
\end{align*}
\begin{IEEEproof}
\begin{align*}
N  \left( x; a, b\right) & N\left( y;  \alpha x, \beta\right)\\
 = & \frac{1}{2 \pi b \beta} \exp \left\{\frac{-\left(x - a\right)^2}{2b^2} + \frac{-\left(y - \alpha x\right)^2}{2 \beta^2}\right\}\\
 = & \frac{1}{2 \pi b \beta} \exp \left\{- \frac{x^2 - 2x \frac{a \beta^2 + \alpha y b^2}{\beta^2 + \alpha^2 b^2} + \frac{a^2 \beta^2 + y^2 b^2}{\beta^2 + \alpha^2 b^2}}{2 \frac{b^2 \beta^2}{\beta^2 + \alpha^2 b^2}}\right\}.
\end{align*}

Then, the result is obtained since the above expression equals
\begin{align*}
\frac{1}{2 \pi b \beta} \exp \left\{- \frac{\left(x - \frac{a \beta^2 + \alpha y b^2}{\beta^2 + \alpha^2 b^2}\right)^2}{2 \frac{b^2 \beta^2}{\beta^2 + \alpha^2 b^2}} - \frac{\left(y- \alpha a\right)^2}{2\left(\beta^2 + \alpha^2 b^2\right)}\right\}
\end{align*}
by completing the squares.
\end{IEEEproof}
\end{lema}

\hspace{0.35cm}\textit{Proof of Theorem \ref{teoremaRecursion}:}

 In the following, $s_{t:t+i}$ and $\V{r}_{t+1:t+i}$ denote the sequences $\left\{s_t, s_{t+1}, ..., s_{t+i}\right\}$ and $\left\{\V{r}_{t+1}, \V{r}_{t+2}, ..., \V{r}_{t+i}\right\}$ respectively, for any $i$.

We proceed by induction, for $i = 1$ we have that
\begin{align}
p& \left(s_{t+1}|s_t, \V{r}_{t+1}\right) \propto p\left(s_{t+1}, s_t, \V{r}_{t+1}\right) \nonumber \\
& = p(\V{r}_{t+1}|s_{t+1}, s_t) p(s_{t+1}|s_t) p(s_t) \nonumber \\
\label{pprop1}
& \propto p\left(\V{r}_{t+1}|s_{t+1}\right) p\left(s_{t+1}|s_{t}\right) \\
\label{pprop2}
 & \propto N\left(s_{t+1};\V{u}_r^{\text{T}} \B{\eta}_{r, c}, \sigma_{r, c}\right) N\left(s_{t+1};\V{u}_s^{\text{T}} \B{\eta}_{s, c}, \sigma_{s, c}\right) 
\end{align}
where proportionalty relationships are due to the fact that $s_t$ and $\V{r}_{t+1}$ are known, \eqref{pprop1} is obtained because the conditional distribution of $\V{r}_{t+1}$ depends only on $s_{t+1}$ since $\{s_{t}, \V{r}_t\}_{t\geq 1}$ form a HMM, and \eqref{pprop2} is obtained because we model conditional distributions as Gaussian given by (\ref{dgauss}) and (\ref{rgauss}). 

Using the previous Lemma, \eqref{pprop2} leads to \eqref{pgauss} with $\hat{s}_{t+1}$ and $\hat{e}_{t+1}$ given by \eqref{spred} and \eqref{epred}, respectively, since $\hat{s}_t = s_t$ and $\hat{e}_t = 0$.

If the statements hold for $i-1$, then for $i$ we have that
\begin{align}
&& p&\left(s_{t+i}|s_t, \V{r}_{t+1:t+i}\right)\propto p\left(s_{t+i}, s_t, \V{r}_{t+1:t+i}\right) \nonumber \\
\label{eq16}
&&= &\displaystyle\int p\left(s_t, s_{t+i-1:t+i}, \V{r}_{t+1:t+i}\right) ds_{t+i-1}\\
\label{eq17}
&&=& \displaystyle\int p\left(s_t, s_{t+i-1:t+i}, \V{r}_{t+1:t+i-1}\right) p\left(\V{r}_{t+i}|s_{t+i}\right) ds_{t+i-1} \\
&& =& \hspace{0.1cm}p\left(\V{r}_{t+i}|s_{t+i}\right) \nonumber \\
\label{eq18}
&&& \cdot \displaystyle\int p\left(s_t, s_{t+i-1}, \V{r}_{t+1:t+i-1}\right) p\left(s_{t+i}|s_{t+i-1}\right) ds_{t+i-1}\\
&& \propto & \hspace{0.1cm} p\left(\V{r}_{t+i}|s_{t+i}\right) \nonumber \\
&&& \cdot \displaystyle\int  p\left(s_{t+i-1}|s_t, \V{r}_{t+1:t+i-1}\right)p\left(s_{t+i}|s_{t+i-1}\right) ds_{t+i-1} \nonumber \\
&& \propto & \hspace{0.1cm} N\left(s_{t+i};\V{u}_r^{\text{T}} \B{\eta}_{r, c}, \sigma_{r, c}\right) \cdot \nonumber\\
\label{eq20}
&&&\hspace{-0.6cm}\displaystyle\int  N\left(s_{t + i -1}; \hat{s}_{t + i -1}, \hat{e}_{t + i -1}\right) N\left(s_{t+i};\V{u}_s^{\text{T}} \B{\eta}_{s, c}, \sigma_{s, c}\right) ds_{t + i -1}
\end{align}
where proportionalty relationships are due to the fact that $s_t$ and $\V{r}_{t+1:t+i}$ are known, \eqref{eq16} is obtained by marginalizing, \eqref{eq17} and \eqref{eq18} are obtained by using the properties of HMMs, and \eqref{eq20} is obtained by using the induction hypothesis and the models of conditional distributions as Gaussians given by (\ref{dgauss}) and (\ref{rgauss}). Then, the result is obtained by applying the previous Lemma to \eqref{eq20} twice, and substituting $\V{u}_s = [1, s_{t+i-1}]^{\text{T}}$.
$\hfill\blacksquare$
\end{appendices}

\bibliographystyle{IEEEtran}
\bibliography{IEEEabrv,bibliography}

\begin{IEEEbiography} 
[{\includegraphics[width=1in, height=1.25in, clip, keepaspectratio]{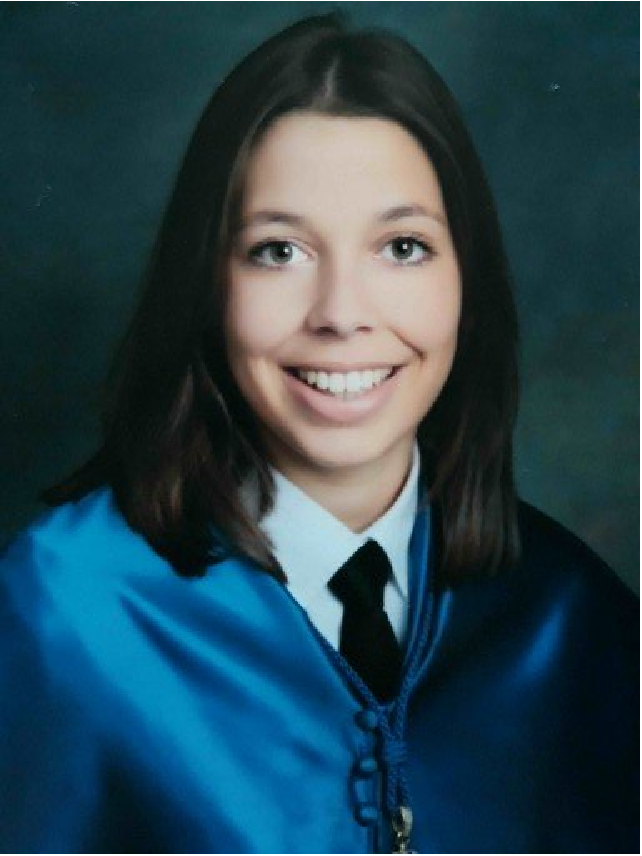}}]{Ver\'onica \'Alvarez} (S'20) received the degree in Mathematics from the University of Salamanca, Spain, in 2019, and her M.Sc degree in Mathematical Research from the Polytechnic University of Valencia, Spain, in 2020.

Since July 2019, she is at the Basque Center for Applied Mathematics (BCAM) where she is currently working towards her Ph.D. degree in collaboration with Iberdrola Innovation. She has experience working on multiple research projects funded by the Spanish ministry of Science and the Basque Government. Her main scientific interests include statistics, data science, and machine learning.
\end{IEEEbiography}

\begin{IEEEbiography} 
[{\includegraphics[width=1in, height=1.25in, clip, keepaspectratio]{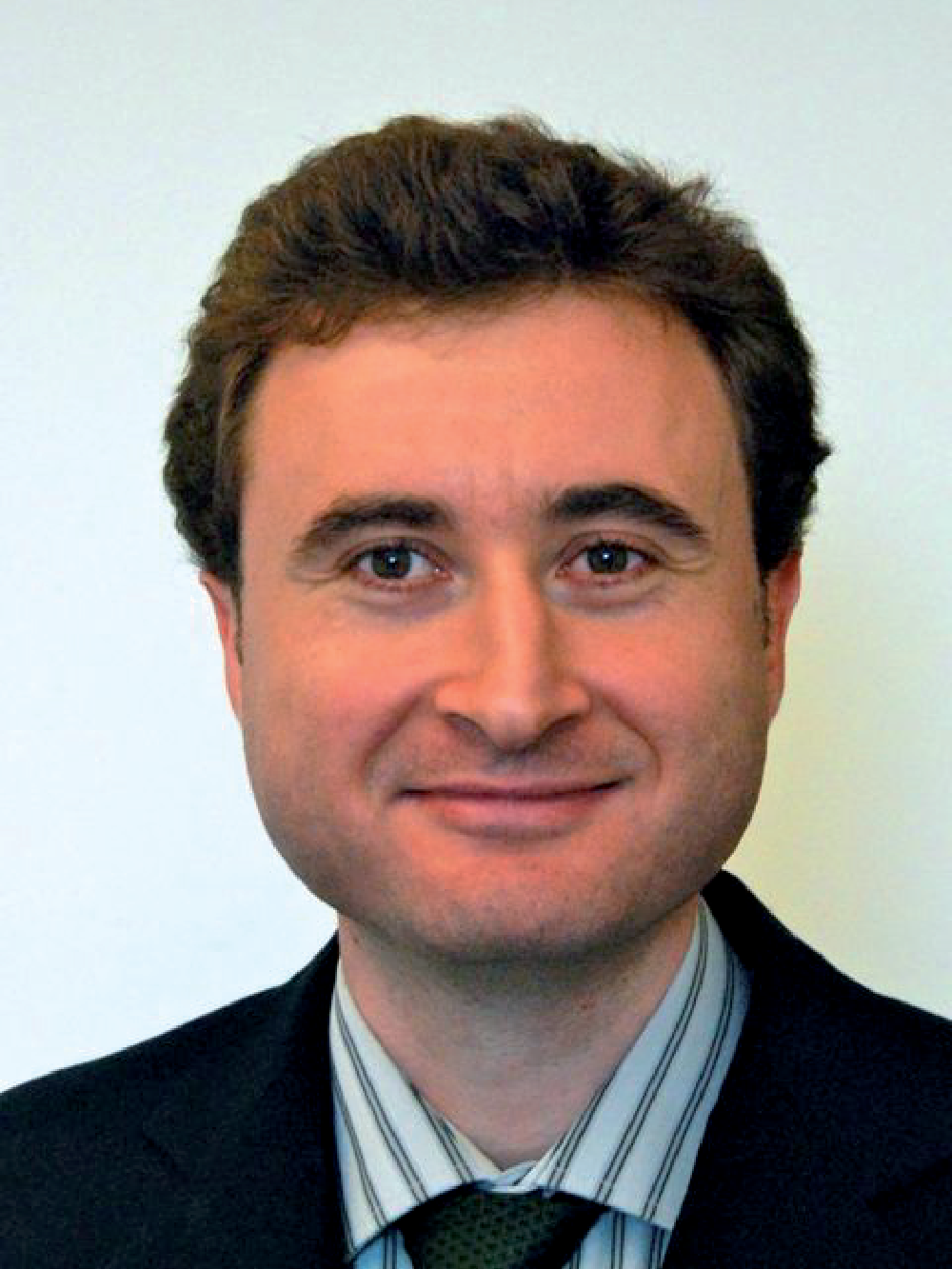}}]{Santiago Mazuelas} (M'10-SM'17) received the Ph.D. in Mathematics and Ph.D. in Telecommunications Engineering from the University of Valladolid, Spain, in 2009 and 2011, respectively. 

Since 2017 he has been Ramon y Cajal Researcher at the Basque Center for Applied Mathematics (BCAM). Prior to joining BCAM, he was a Staff Engineer at Qualcomm Corporate Research and Development from 2014 to 2017. He previously worked from 2009 to 2014 as Postdoctoral Fellow and Associate in the Wireless Information and Network Sciences Laboratory at the Massachusetts Institute of Technology (MIT). His general research interest is the application of mathematics to solve practical problems, currently his work is primarily focused on statistical signal processing, machine learning, and data science. 

Dr. Mazuelas is Area Editor (signal processing) for the IEEE Communications Letters since Oct. 2019 (Associate Editor from Jan. 2017 to Sep. 2019), and served as Technical Program Vice-chair at the 2021 IEEE Globecom as well as Symposium Co-chair at the 2014 IEEE Globecom, the 2015 IEEE ICC, and the 2020 IEEE ICC. He has received the Young Scientist Prize from the Union Radio-Scientifique Internationale (URSI) Symposium in 2007, and the Early Achievement Award from the IEEE ComSoc in 2018. His papers received the IEEE Communications Society Fred W. Ellersick Prize in 2012, and Best Paper Awards from the IEEE ICC in 2013, the IEEE ICUWB in 2011, and the IEEE Globecom in 2011.
\end{IEEEbiography} 

\begin{IEEEbiography}
[{\includegraphics[width=1in, height=1.25in, clip, keepaspectratio]{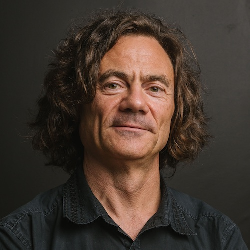}}]{Jos\'e. A. Lozano} (M'04-SM'19-F'21) received the B.S. degree in Mathematics and the B.S. degree in Computer Science from the University of the Basque Country, San Sebastian Donostia, Spain, in 1991 and 1992, respectively, and the Ph.D. degree in Computer Science from the University of the Basque Country in 1998. He is scientific director of the Basque Center for Applied Mathematics and full professor in Computer Science and Artificial Intelligence at the University of the Basque Country UPV/EHU. His research interests focus on machine learning, heuristic optimization and its application to different scenarios in biology, medicine and ecology, to name a few. 

Jos\'e A. Lozano is a Fellow of the IEEE and member of the editorial board of the main journals of his scientific field such as the IEEE Trans. on Neural Networks and Learning Systems and IEEE Trans. on Evolutionary Computation. In addition he has contributed to the most important congress of the field being the general chair of  IEEE Congress on Evolutionary Computation 2017 and the European Conference on Machine Learning/Principles and Practice of Knowledge Discovery in Databases ECML/PKDD 2021.
\end{IEEEbiography} 

\end{document}